\documentclass[11pt]{article}
\usepackage[]{acl}

\usepackage{paralist}
\usepackage{times}
\usepackage{latexsym}
\usepackage[T1]{fontenc}
\usepackage[utf8]{inputenc}
\usepackage{microtype}
\usepackage{inconsolata}
\usepackage{graphicx}

\usepackage{booktabs}
\usepackage{array}
\usepackage{multirow}
\usepackage{float}
\usepackage{xcolor}
\usepackage{pifont}
\usepackage{xspace}
\usepackage{enumitem}
\usepackage[most]{tcolorbox}
\usepackage{cleveref}
\usepackage{amsmath}
\usepackage{amssymb}
\usepackage{amsthm}

\newtheorem{definition}{Definition}

\newcommand{\svcomp}{\textsc{SV-Comp}\xspace}  

\title{LLMs versus the Halting Problem: \\ Characterizing Program Termination Reasoning}

\author{
\textbf{Oren Sultan}$^{1,2}$\thanks{Work done during an internship at Meta FAIR.
\\ $^{\dagger}$Joint supervision.},
\textbf{Jordi Armengol-Estapé}$^{1}$,
\textbf{Pascal Kesseli}$^{1}$,
\textbf{Julien Vanegue}$^{3,4}$ \\
\textbf{Dafna Shahaf}$^{2}$,
\textbf{Yossi Adi}$^{1\dagger}$,
\textbf{Peter O'Hearn}$^{1,5\dagger}$\\
\\
$^{1}$Meta FAIR \\
$^{2}$The Hebrew University of Jerusalem, Israel \\
$^{3}$Bloomberg, New York, USA \\
$^{4}$Imperial College London, UK \\
$^{5}$University College London, UK \\
}

\begin{document}
\maketitle
\begin{abstract}

Determining whether a program terminates is a central problem in computer science. Turing’s Halting Problem established termination as undecidable, showing that no algorithm can universally determine termination for all programs and inputs.
Hence, verification tools approximate termination, sometimes failing to prove or disprove; these tools rely on problem-specific architectures and are usually tied to particular programming languages. Recent advances in LLMs raise a natural question: \emph{To what extent can they reason about program termination?}

We evaluate frontier LLMs on a diverse set of $C$ programs from the International Competition on Software Verification (\svcomp) 2025. Our results show that GPT-5 and Claude Sonnet-4.5 achieve scores comparable to top-ranked verification tools (with test-time scaling). However, while models often correctly infer whether programs terminate, they frequently fail to construct a \emph{witness} as formal proof, revealing a gap between semantic recognition and symbolic proof generation. Performance further degrades as code length increases. Beyond witness automaton graphs, we introduce a divergence-precondition formulation that characterizes non-termination conditions as logical constraints.
We hope these findings motivate future research on real-world termination benchmarks, neuro-symbolic approaches that combine LLMs with symbolic verification methods, and, more broadly, LLM reasoning on other undecidable problems.

\end{abstract}

\section{Introduction}

\begin{figure*}[t]
\centering
\includegraphics[width=0.8\textwidth]{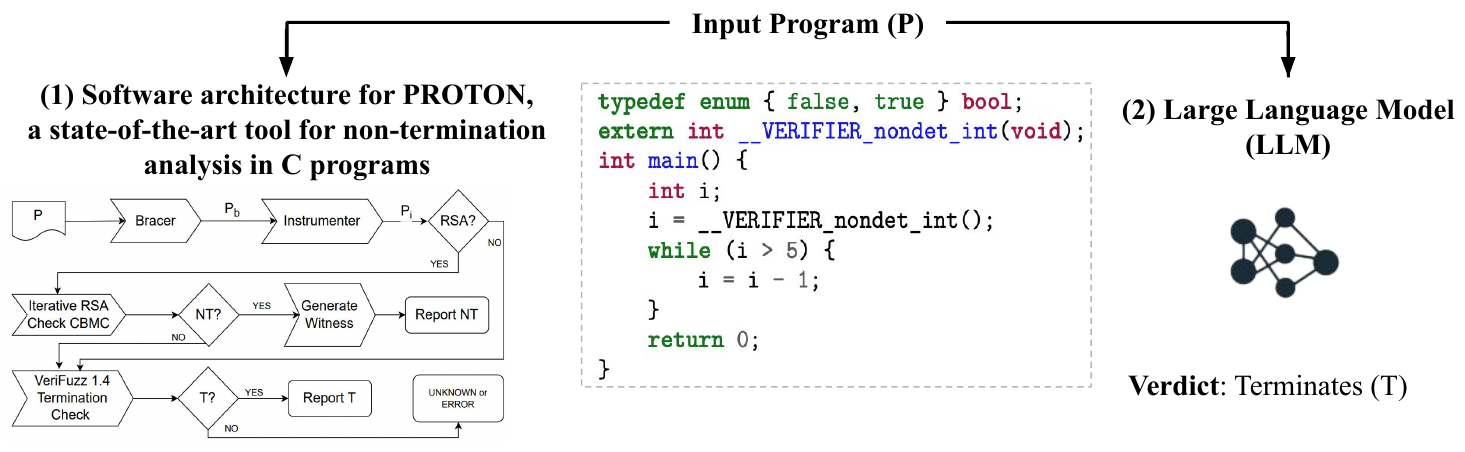}
\caption{
\textbf{Termination prediction task.}
Given a program, the task is to predict whether the program halts for all inputs (instantiations of nondeterministic variables).
\textbf{(1)} Traditional verification tools like PROTON use multi-component architectures for parsing, input augmentation, and tool-chain management (its architecture image reproduced from~\citet{metta2024proton} under \href{https://creativecommons.org/licenses/by/4.0/}{Attribution 4.0 International License}).
\textbf{(2)} We investigate whether LLMs can match state-of-the-art verification tools, providing a simpler, language-agnostic solution, as illustrated in the example above where the LLM correctly predicts termination.
For an example of non-termination see Figure~\ref{fig:witness_example}.
}
\label{fig:halteval_task}
\end{figure*}

\begin{figure*}[t]
\centering
\includegraphics[width=0.85\textwidth]{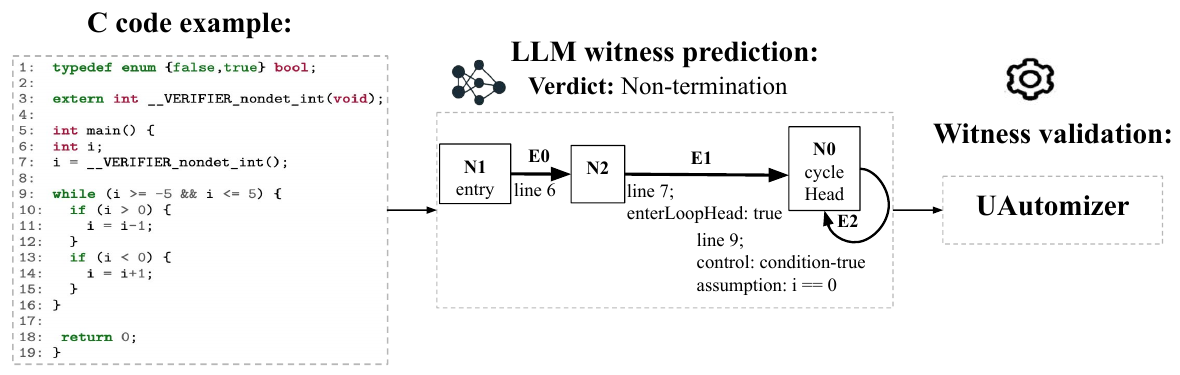}
\caption{
\textbf{LLM non-termination witness prediction.}
Given a $C$ program, an LLM that predicts non-termination must additionally output a \textbf{witness automaton} as a proof in JSON format  (see Figure~\ref{appendix:llm_witness_json}).
The witness automaton models a potentially infinite execution: nodes correspond to program states and edges to transitions.
The predicted JSON is converted to GraphML and validated by a \textbf{witness validator} (e.g., UAutomizer).
The example illustrates a loop where $i$ is initialized in $[-5,5]$, eventually reaches $0$ (see the \emph{assumption} on edge E2), and executes indefinitely.
}
\label{fig:witness_example}
\end{figure*}


Program termination is a central problem in computer science, demonstrated by the Halting Problem~\citep{turing1936computable}. The undecidability of the Halting Problem proves that no algorithm can universally determine whether a given program will halt. 
Ensuring program termination is 
important
for software reliability and safety~\citep{avizienis2004basic}. Non-termination may present a critical software defect that can lead to unbounded resource consumption, loss of system responsiveness, and severe failures in real-world deployments. 


Despite the undecidability of the Halting Problem, a variety of verification tools have been proposed over the years to approximate termination analysis. State-of-the-art  tools, such as UAutomizer~\citep{heizmann2023ultimate} and PROTON~\citep{metta2024proton}, use automata over program statements or leverage 
symbolic execution via
CBMC~\citep{kroening2014cbmc}.
These tools often rely on complex, multi-stage  problem- and 
language-specific architectures (for example, see PROTON architecture in  Figure~\ref{fig:halteval_task}). 

Large Language Models (LLMs) have recently shown strong capabilities in code generation and reasoning~\citep{zheng2023survey,zhao2023survey,liu2024deepseek,jiang2024survey}.
Unlike traditional tools, LLMs can be integrated directly into the coding process, reason beyond surface-level syntax, and generalize across languages. 
We ask: \emph{To what extent can frontier LLMs correctly predict whether a program terminates?} To answer this, we evaluate frontier LLMs on full \svcomp termination tasks under machine-validated proof constraints.

Program termination is undecidable: proving it requires ruling out all infinite execution paths, not merely observing finite traces. Assessing termination requires LLMs to reason about program structure and control flow beyond pattern matching.

In this work, we assess the reasoning capabilities of LLMs on the International Competition on Software Verification (\svcomp) 2025 Termination category\footnote{The dataset is publicly available under the CC-BY-4.0.}~\citep{beyer2012competition}. We cover both open-weights models, i.e., CWM~\citep{carbonneaux2025cwm} and Qwen3-32B~\citep{yang2025qwen3} and proprietary models, i.e., GPT-5~\citep{gpt-5}, Claude Sonnet 4.5~\cite{claude4.5}, and GPT-4o~\citep{hurst2024gpt}, which serves as a non-reasoning baseline model. For a reference, we report results for the top three \svcomp 2025 verification tools: PROTON~\citep{metta2024proton}, UAutomizer~\citep{heizmann2023ultimate}, and AProVE~\citep{emrich2023aprove}. 


Results indicate that GPT-5 and Claude Sonnet-4.5 trail only PROTON, the first-place tool. CWM ranks immediately behind UAutomizer, which finished second.
Error analysis reveals a substantial classification-proof gap: while LLMs are often highly effective at predicting program termination, they frequently fail to construct verifier-valid witness automata for infinite execution paths. Additionally, their performance tends to degrade when handling longer, more complex code samples.

To further validate our findings, we analyze a subset of programs by prompting the model to specify a {\em precondition}\/ (a logical formula) describing the input values that cause non-termination. This is related to the concept of weakest precondition from program verification \citep{Dijkstra1976}.
We believe that such divergence preconditions are simpler and more interpretable than \svcomp witnesses.

Note that our goal is to evaluate current LLMs on program termination prediction, rather than optimize them for this task, propose a new termination-analysis algorithm, or design a new evaluation protocol. We intentionally adopt the existing, community-standard \svcomp task definition, label space, witness format, and scoring scheme, since this enables reproducible evaluation and direct comparison against state-of-the-art termination analyzers. Our contribution is therefore best understood as a systematic empirical characterization of LLM termination reasoning under this established protocol, rather than a new benchmark or methodology; we discuss more realistic, LLM-specific termination benchmarks as an important direction for future work (Section~\ref{sec:conclusions_and_future_work}). We believe that understanding these strengths and limitations can shed light on the broader capabilities of LLMs for challenging computational reasoning tasks. We view program termination as a useful testbed for studying frontier LLM reasoning on undecidable semantic tasks, motivating future research on real-world termination benchmarks, neuro-symbolic systems that combine LLMs with symbolic verifiers, and reasoning about additional undecidable properties.

Termination is one instance of a large class of undecidable semantic program properties. By Rice’s Theorem~\citep{rice1953classes}, every non-trivial semantic property of programs is undecidable, including central verification tasks such as unreachability, memory safety, and program equivalence.

\textbf{Our contributions are:}
\textbf{(1)} We present the first large-scale study of frontier LLM reasoning on full \svcomp termination tasks, providing a new testbed for studying semantic reasoning on undecidable problems.
\textbf{(2)} We show that frontier reasoning models approach specialized symbolic analyzers in prediction accuracy, while revealing a substantial gap between correct semantic prediction and valid symbolic proof construction. \textbf{(3)} We identify key failure modes of LLM reasoning: witness generation remains the primary bottleneck, performance degrades with code length, and errors concentrate in concurrency over standard control flow.
\textbf{(4)} We introduce a domain witness formulation for non-termination, enabling a simpler and more interpretable analysis of model reasoning.
\textbf{(5)} We release data and code at: \url{https://github.com/orensul/LLMs_vs_Halting_Problem}.

\section{Problem Formulation}
\label{sec:problem_formulation}

We follow the format specification of the \svcomp  Termination category, in which the \textbf{input} is a $C$ program with one main function and possibly auxiliary helper functions. The program may include multiple non-deterministic variables, which can be assigned random values during execution. The \textbf{output} is a termination prediction. 
Specifically, we consider three possible predictions: \textbf{(1) T (Terminating)}, meaning the program terminates on all possible input values; \textbf{(2) NT (Non-Terminating)}, indicating the program diverges on some values and forms a non-termination loop. Here, the model is required to provide a \emph{witness automaton}, which is a graph containing a cycle of valid states to demonstrate an infinite execution path within the program. Nodes correspond to program states; edges represent transitions. 
Importantly, this path does not need to be fully explicit; the model is not required to specify the exact nondeterministic values chosen along the path. As a result, the model does not need to produce a concrete counterexample, but rather a valid infinite path through the program's state space. This allows standard \svcomp validators to confirm correctness in under a second.
See Appendix~\ref{appendix:UAutomizer} for details about UAutomizer, the witness validator we used, and Figure~\ref{fig:witness_example} for an example of LLM witness prediction.
\textbf{(3) UNK (Unknown)} is returned when the model cannot determine the termination behavior.

\section{Dataset}
\label{sec:dataset}

We use \svcomp 2025, which is arguably the largest public collection of formal verification tasks, spanning six key properties for tool evaluation. 
We focus on the \texttt{termination} category, comprising a total of $2{,}328$ tasks.
These tasks are organized into four subcategories: BitVectors (bit-precise arithmetic), MainControlFlow (complex control flow such as loops and recursion), MainHeap (dynamic heap memory manipulation), and Other.
Each subcategory presents distinct program analysis challenges. A detailed summary of these subcategories is provided in Table~\ref{tab:termination_category_summary} in the Appendix. Each task in the dataset provides a $C$ code and its expected verification result, i.e., whether the code terminates or not. The benchmarks range from synthetic cases targeting specific algorithms to real-world code. Figure~\ref{fig:witness_example} illustrates a non-termination sample from the \svcomp dataset. 


\noindent\textbf{Dataset statistics.} The dataset contains $C$ programs of diverse lengths. Using the Instruct Llama3 tokenizer, we find that $90$\% of samples contain at most $20$K tokens (median: $6$K). See Appendix~\ref{appendix:dataset_stats} for details.
The dataset contains $2{,}328$ samples, $65.6$\% labeled as Termination (T) and $34.4$\% as Non-termination (NT). This distribution is skewed toward terminating programs, reflecting typical real-world software; non-terminating cases often arise from unintended behaviors or bugs, which are also deliberately included in \svcomp.

\section{Experimental Setup}

\begin{figure*}[t]
\includegraphics[width=0.82\textwidth]{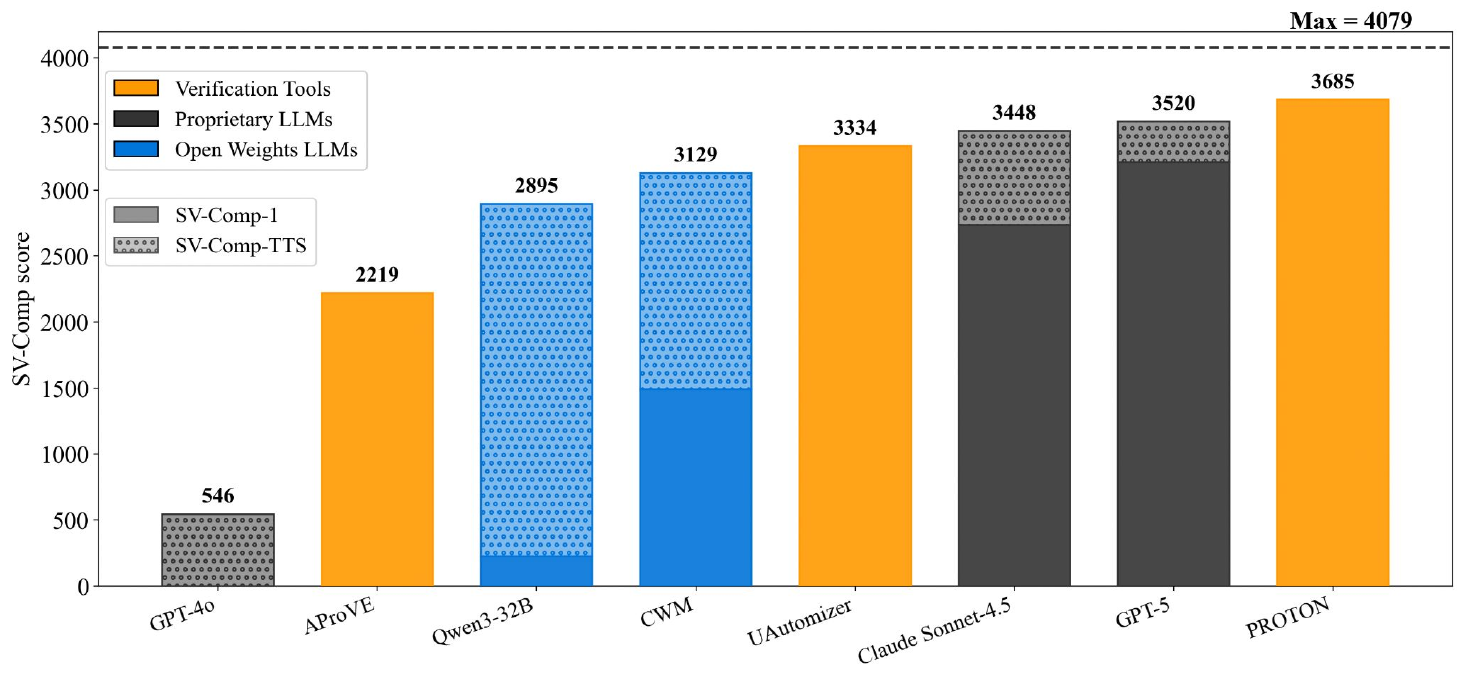}
\centering
    
\caption{
\textbf{\svcomp main results.}
Mean \svcomp scores for LLMs (across $100$ bootstraps), top \svcomp 2025 verification tools, and the max possible scores (\textbf{minimum score is $-50{,}064$}). GPT-5 (TTS) and Claude Sonnet-4.5 (TTS) would place 2nd and 3rd ($3{,}520$ and $3{,}448$), behind the gold medalist PROTON. CWM ranks just below UAutomizer. GPT-4o scores significantly lower: $546$ with TTS, and a negative score without TTS ($-5,145$).
\label{fig:svcomp_main_results}
}
\end{figure*}

We provide a detailed description of the evaluation setup, covering model inference, evaluation methodology, and task-specific evaluation metrics. 

\subsection{Model Inference}

We evaluate CWM~\cite{carbonneaux2025cwm}, Qwen3-32B~\cite{yang2025qwen3}, GPT-5~\cite{gpt-5} and Claude Sonnet-4.5~\cite{claude4.5}, all in reasoning mode. We additionally evaluate GPT-4o~\cite{hurst2024gpt} as a strong baseline model with no reasoning capabilities.


\noindent\textbf{Inference pipeline.} Each model is prompted to generate a JSON object containing a \texttt{verdict} attribute: \texttt{true} for Termination (T), and \texttt{false} for Non-termination (NT). For NT cases, the model also outputs a \texttt{witness} attribute, which is a JSON graph (nodes are states in the program; edges the transitions between states) representing a potential infinite path. We use the same prompt across all models to ensure a controlled comparison. We deliberately avoid task-specific tuning (e.g., fine-tuning or prompt engineering) to reflect the models' general performance. Despite the lack of task-specific tuning, models still achieve strong performance.
See Figure~\ref{appendix:svcomp_prompt} in the Appendix for the complete prompt provided to the model.

For a false \emph{verdict} (NT), the LLM outputs a witness automaton graph in the \emph{witness} JSON attribute. Predictions are schema-validated: nodes and edges must have unique ids and include required attributes (\texttt{id}, \texttt{source}, \texttt{target}, \texttt{line}\footnote{We add code line numbers supporting \texttt{line} prediction.}, \texttt{sourcecode}).
Duplicates or missing fields are treated as errors.
These ensure that outputs are syntactically valid and schema-compliant before validation; most failures arise during downstream semantic validation rather than at this stage (see Section~\ref{subsec:errors_analysis}). See Figure~\ref{appendix:llm_witness_json} in Appendix for an example.
Finally, we convert the validated JSON object into a GraphML, which is then processed by a witness validator. We use the UAutomizer ~\cite{heizmann2023ultimate}, which symbolically checks the feasibility of the candidate infinite execution using an SMT solver. See Appendix~\ref{appendix:UAutomizer} for details.

\subsection{\svcomp Score}
The \svcomp score averages performance across $k$ categories with equal weight:
$Score = \frac{1}{k} \sum_{i=1}^{k} \frac{s_i}{n_i} \left(\sum_{i=1}^{k} n_i\right)$,
where $s_i$ and $n_i$ are the total score and number of samples in category $i$.

\noindent\textbf{Scoring per sample.} We use asymmetric scoring, following \svcomp, with non-termination (NT) as the positive class. See Table~\ref{tab:scoring_per_sample} for the scoring criteria.
This scheme prioritizes safety: FNs (misclassifying diverging programs as terminating) may lead to deploying software with infinite loops, risking system failures and resource exhaustion.

\begin{table}[!t]
    \centering
    \footnotesize
    \setlength{\tabcolsep}{10pt}
    \resizebox{\columnwidth}{!}{%
    \begin{tabular}{l | c}
    \toprule
    \textbf{Prediction type} & \textbf{Score} \\
    \midrule
    Correct Termination (TN) & +2 \\ 
    Correct Non-termination (TP\_valid\_witness) & +1 \\  
    \midrule
    Unknown & 0 \\
    Correct Non-termination (TP\_invalid\_witness) & 0 \\
    \midrule 
    Incorrect Non-termination (FP) & -16 \\
    Incorrect Termination (FN) & -32 \\
    \bottomrule
    \end{tabular}}
  \caption{\textbf{\svcomp scoring per sample.} Non-termination (NT) is the positive class. For NT predictions, scores differ based on witness validity.}
    \label{tab:scoring_per_sample}
\end{table}

\subsection{Evaluation}
\noindent\textbf{Bootstrap sampling.}
LLM outputs are inherently stochastic, so repeated runs may yield different predictions. To ensure a robust evaluation, we employ bootstrap sampling. For each instance, we aggregate results across $100$ bootstrap runs.

\noindent\textbf{\svcomp-1.} We generate $20$ predictions per instance and randomly select one prediction from the $20$ generated for each bootstrap iteration.

\noindent\textbf{\svcomp-TTS.}
To improve reliability and uncertainty estimates, we apply Test-Time Scaling (TTS) with Consensus Voting. For each instance, we randomly select $n=10$ out of $20$ predictions. If all sampled votes agree (considering \texttt{T} and \texttt{NT}, while ignoring \texttt{unknown}), we adopt the unanimous prediction. If there is any disagreement, we output \texttt{unknown} to reflect model uncertainty, reducing the risk of penalties from incorrect predictions.

We found $n=10$ as a sweet spot for TTS, representing an optimal trade-off. Lower values of $n$ make consensus easier and reduce \texttt{unknown} outputs, while higher values increase the likelihood of disagreement and thus more \texttt{unknown} predictions, which can lower the overall score.

Note that requiring agreement between all trajectories is a strict criterion that may lead to excessive \texttt{unknown} predictions and reduced overall score. Future work could explore alternative agreement thresholds to optimize performance.

\noindent\textbf{F1 score.} Beyond the \svcomp score, we report per-class F1 scores to enable a more fine-grained analysis of model performance. Unknown predictions are considered mistakes.

\noindent\textbf{Model hyperparameters.} We configured each model using its official recommended settings. Top-p was set to $0.95$.
Temperatures are set to 1.0 (CWM, Claude), 0.6 (Qwen), and 0.7 (GPT-4o). GPT-5 does not support temperature; we use the default configuration with \texttt{medium} reasoning effort.


\noindent\textbf{Compute considerations.}
Proprietary models are accessed via API, while open-weight models run on our internal GPUs under a fixed budget. We rely on official \svcomp~2025 results~\cite{beyer2025improvements}, evaluated under fixed limits (15 min CPU, 4 cores, 15 GB RAM). Note that GPU/API inference is generally more expensive than the CPU-based alternatives; we focus on \emph{absolute} performance rather than cost-normalized comparisons. The LLM and \svcomp tool scores are therefore obtained under different computational conditions and are not directly cost-comparable; moreover, \svcomp-TTS produces substantially more generations per instance than naive single-shot prediction. To make LLM inference cost concrete, we report LLM token consumption in Appendix~\ref{appendix:compute_cost}.

\label{subsec:setup}

\section{Results}
\label{subsec:results}

\subsection{Main Results}
\label{subsec:main_results}

Results are presented in Figure~\ref{fig:svcomp_main_results}. Under the TTS setup, all reasoning models show impressive \svcomp scores. GPT-5 ($3{,}520$) and Claude Sonnet-4.5 ($3{,}448$) achieve scores in the range of the top \svcomp tools, trailing only PROTON.
CWM ($3{,}129$) and Qwen3-32B ($2{,}895$) follow, and do not exceed UAutomizer. GPT-4o, serving as a non-reasoning baseline, shows substantially lower performance. While all models benefit from TTS, this benefit is more pronounced in CWM, Qwen3-32B, and GPT4o (e.g., Qwen \svcomp-1 is $225$ while its \svcomp-TTS is $2{,}895$). Interestingly, the gap is the smallest for GPT-5, implying better model calibration and robustness across predictions. 

Beyond the main \svcomp score, Table~\ref{tab:svcomp_f1_scores} reports per-class F1 scores for both T and NT. The F1 ranking closely matches the \svcomp ranking, with GPT-5 performing best overall. F1 scores decrease under TTS because additional \texttt{unknown} predictions are neutral in \svcomp scoring but treated as errors in F1. Standard deviations remain consistently low ($<0.1$), indicating stable performance across runs. 
See Appendix~\ref{appendix:main_results} for details.


\begin{table}[!t]
    \centering
    \footnotesize
    \setlength{\tabcolsep}{6pt}
    \begin{tabular}{l | c c}
        \toprule
        \textbf{Model} & \textbf{F1 (T) ~$\uparrow$} & \textbf{F1 (NT) ~$\uparrow$} \\
        \midrule
        GPT-5 (TTS)         & \textbf{0.98} & 0.96 \\
        GPT-5               & \textbf{0.98} & \textbf{0.97} \\
        \midrule
        Claude S-4.5 (TTS)  & 0.94          & 0.89 \\
        Claude S-4.5        & \textbf{0.98} & 0.96 \\
         \midrule
        CWM (TTS)           & 0.84          & 0.63 \\
        CWM                 & 0.91          & 0.80 \\
         \midrule
        Qwen3-32B (TTS)     & 0.83          & 0.59 \\
        Qwen3-32B           & 0.90          & 0.78 \\
         \midrule
        GPT-4o (TTS)        & 0.73          & 0.50 \\
        GPT-4o              & 0.73          & 0.49 \\ 
        \bottomrule
    \end{tabular}
\caption{
    \textbf{LLMs F1 scores per class.} 
    F1 (T) and F1 (NT) are averaged over 100 bootstrap runs. F1 decreases under TTS due to increased \texttt{unknown} predictions (counted as errors in F1). Rankings align with \svcomp score.
}
    \label{tab:svcomp_f1_scores}
\end{table}

\subsection{Analysis}
\label{subsec:errors_analysis}

Next, we evaluate models along four complementary axes, each targeting a distinct aspect of termination reasoning beyond overall \svcomp score: \textbf{(1) witness validity}, which measures whether models can back a prediction with formally verifiable evidence, not just a correct label; \textbf{(2) Unknown rate (\svcomp-1, TTS)}, which measures calibration, i.e., whether models abstain appropriately rather than guess; \textbf{(3) performance vs. code length}, using length as a proxy for reasoning complexity; and \textbf{(4) performance across \svcomp subcategories}, which identifies which classes of termination problems remain most challenging.

\begin{table}[!t]
    \centering
    \footnotesize
    \setlength{\tabcolsep}{10pt}
    \begin{tabular}{l | c | c | c }
    \toprule
    \textbf{Model} &  \textbf{P} ~$\uparrow$ &  \textbf{R} ~$\uparrow$ &   \textbf{V} ~$\uparrow$  \\
    \midrule
    GPT-5 & \textbf{40.1\%} & \textbf{40.7\%} & \textbf{41.4\%}  \\
    Claude Sonnet-4.5& 38.5\% & 39.9\% & 40.7\%   \\
    CWM & 28.5\% & 25.9\% & 30.6\%   \\
    Qwen3-32B & 22.6\% & 17.9\% & 24.8\%   \\
    GPT-4o & 20.8\% & 22.7\% & 27.3\%   \\
    \bottomrule
    \end{tabular}
  \caption{\textbf{LLMs witness prediction  (validated by UAutomizer) success rates}. Metrics: \textbf{P} (Precision): (\%) of validated witness NT predictions out of all NT predictions; \textbf{R} (Recall): (\%) of validated witness NT predictions out of all NT samples; \textbf{V} (Validity): (\%) of validated witness NT predictions out of all \texttt{correct} NT predictions. GPT-5 and Claude lead in prediction rates, while CWM, GPT-4o, and Qwen3-32B perform lower, highlighting the challenge of generating a valid witness.
    }
    \label{tab:errors_analysis_witness}
\end{table}

\begin{table}[t!]
    \centering
    \footnotesize
    \setlength{\tabcolsep}{10pt}
    \begin{tabular}{l | c | c}
    \toprule
    \textbf{Model} & \textbf{Unk} ~$\downarrow$ & \textbf{TTS-Unk} ~$\downarrow$\\
    \midrule
    GPT-5   & 0.5\% & \textbf{3\%} \\
    Claude Sonnet-4.5 & \textbf{0.3\%} & 8\% \\
    CWM     & 6\%   & 22\% \\
    Qwen3-32B   & 2\%   & 23\% \\
    GPT-4o  & 26\%  & 30\% \\
    \bottomrule
    \end{tabular}
    \caption{\textbf{LLMs' \texttt{unknown} prediction distribution.}
\textbf{Unk}: (\%) of \texttt{unknown} predictions, aggregated over 20 generations per entry. \textbf{TTS-Unk}: (\%) of cases without unanimous agreement among 10 model predictions, resulting in an \texttt{unknown}. GPT-5 and Claude Sonnet-4.5 rarely predict \texttt{unknown}, while GPT-4o does so most often. For TTS-Unk, GPT-5 and Claude show high prediction unanimity, whereas CWM, Qwen3-32B, and GPT-4o show more disagreement, improving \svcomp scores but lowering F1 scores.
}

    \label{tab:errors_analysis_unk}
\end{table}

\noindent\textbf{Witness automaton prediction.} Table~\ref{tab:errors_analysis_witness} summarizes the results of witness automaton prediction, as validated by the UAutomizer witness validator. Performance is evaluated using three metrics, which correspond to the prediction types defined in Table~\ref{tab:scoring_per_sample}.
The first metric, Validity (V), measures the proportion of correct NT predictions that have a valid witness, relative to all \emph{correct} NT predictions:
Validity $= \frac{TP_{\text{valid}}}{TP_{\text{valid}} + TP_{\text{invalid}}}$.
Similarly, Precision (P) and Recall (R) are defined as the proportion of correct NT predictions with a valid witness (out of all NT predictions in precision, and out of all NT samples in recall).
Note that Precision and Recall also account for NT label prediction errors: false positives (FP) in Precision and false negatives (FN) in Recall, which are independent of whether the witness is valid. GPT-5 achieves the highest witness automaton validated predictions, with Claude performing similarly. CWM, GPT-4o, and Qwen3-32B exhibit lower performance, with CWM providing the best performance among the three.





To analyze failure modes, we distinguish between \emph{formatting} errors (e.g., invalid schema or graph structure) and \emph{content} errors. Formatting failures are relatively rare: GPT-5 and GPT-4o produce none, Qwen and Claude around $5\%$, and CWM around $25\%$. Most outputs successfully pass syntax validation, and failures are primarily content-related: witnesses are well-formed but rejected by the validator as semantically invalid.

This reveals a gap between semantic prediction and proof construction: models often predict non-termination correctly yet fail to produce valid witnesses. Similar gaps have been observed in theorem proving and mathematical reasoning, where models generate correct answers more reliably than verifier-valid derivations~\citep{sultan2025towards}. Motivated by this gap, we introduce an alternative witness format in Section~\ref{subsec:further_results}.

\noindent\textbf{Unknown prediction.} Table~\ref{tab:errors_analysis_unk} presents the results for unknown prediction considering both \svcomp-1 and TTS. GPT-5 and Claude predict \texttt{unknown} least often, $0.5$\%, $0.3$\% respectively, followed by Qwen3-32B ($2$\%) and CWM ($6$\%). CWM’s higher \texttt{unknown} rate improves its \svcomp-1 score compared to Qwen. GPT-4o predicts \texttt{unknown}  much more frequently ($26$\%), reflecting its difficulty with the classification task.

Under the TTS setup, GPT-5 and Claude have the lowest \emph{unknown} rates ($3$\% and $8$\%). These results are due to better consensus among model predictions, which implies better and more consistent model predictions. CWM, Qwen3-32B, and GPT-4o show less agreement, resulting in higher \emph{unknown} rates, $22$\%, $23$\%, $30$\%, respectively.


\begin{figure}[t!]
\includegraphics[width=0.5\textwidth]{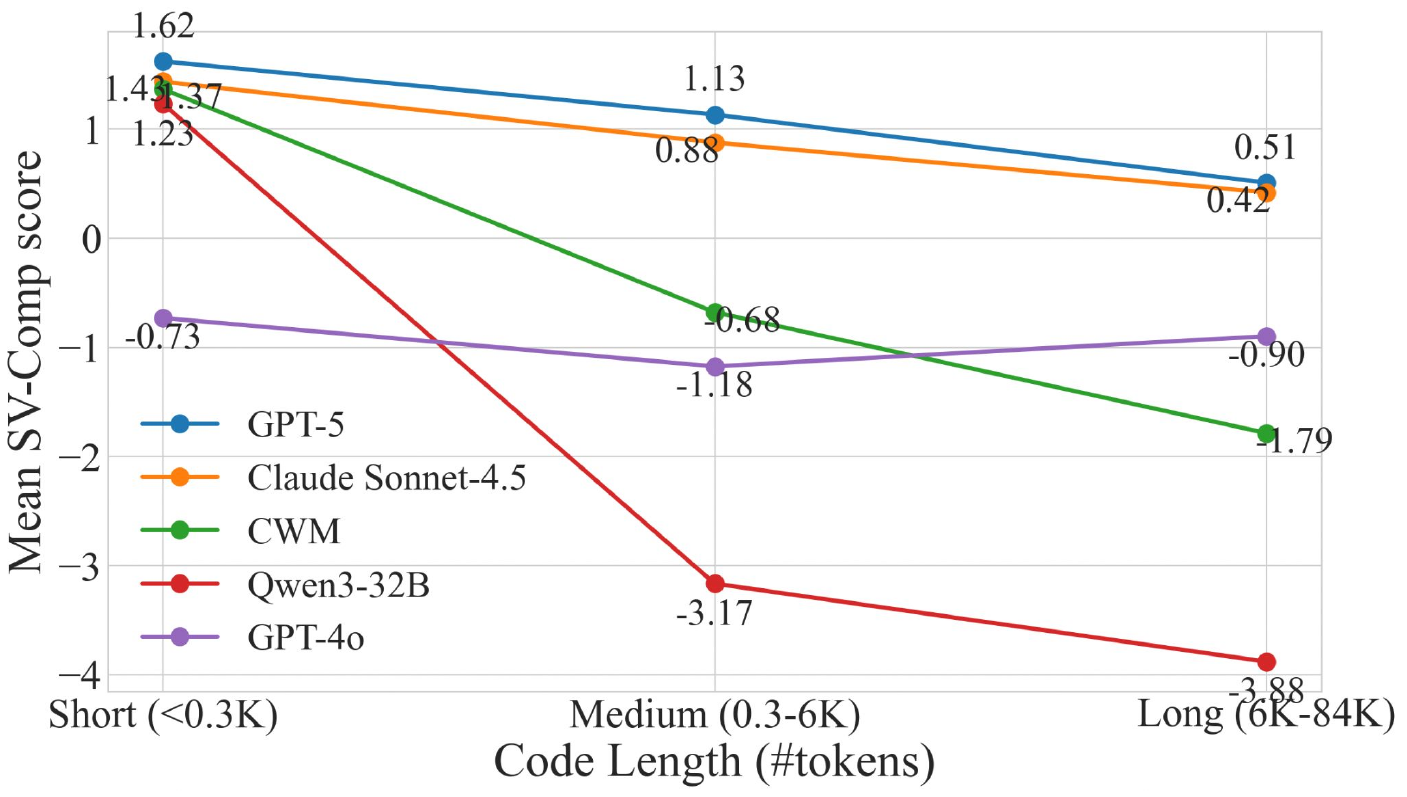}
\centering
\caption{
    \textbf{LLMs mean \svcomp score vs. code length}.
    Dataset examples are grouped into three \emph{equal size} bins by code length (measured in tokens using Instruct Llama3 Tokenizer) (\textbf{x} axis). 
    The mean \svcomp score (\textbf{y} axis) per bin is shown for all models (20 predictions per sample). Scores decrease with code length; GPT-5 leads, followed by Claude and CWM. }
\label{fig:code_length_bins}
\end{figure}

\noindent\textbf{Code input length.}
We analyze performance as a function of code complexity, using input length as a proxy (longer programs are typically more challenging).
Figure~\ref{fig:code_length_bins} shows the mean \svcomp score across three equal-size bins by input length (measured in tokens using the Instruct Llama3 tokenizer). Performance consistently decreases with length.
GPT-5 achieves the highest scores across all bins, followed by Claude and CWM. In the medium and long bins, GPT-4o outperforms Qwen3-32B and matches CWM’s performance, although its scores are notably lower in the smallest bin. GPT-4o predicts a relatively high number of \texttt{unknown} outcomes, which helps reduce penalties for longer code inputs.
Since code length is only a proxy for complexity (e.g., number of variables and control-flow depth), we do not claim it is causal. Rather, we report an empirical degradation consistent with similar length effects in multi-step mathematical reasoning. One possible explanation is that longer programs require globally consistent reasoning over richer control flow and larger state spaces.

\noindent\textbf{Sub-category error rates.}
We analyze GPT-5, the top-performing model, and find that errors concentrate in multi-threaded programs, while simpler control-flow categories (e.g., loops, arrays, recursion) show near-zero failure rates. We report this as an empirical association rather than a causal explanation for why concurrency is harder, which we leave to future work. See Appendix~\ref{appendix:subcat_errors} for details.

\noindent\textbf{Evidence against data memorization.}
Several findings suggest that the results are unlikely to be explained by benchmark memorization. \textbf{(1)} Models recover non-termination domains in a novel witness format not available online (Section~\ref{subsec:further_results}); \textbf{(2)} performance degrades with code length and complexity; \textbf{(3)} non-reasoning models perform substantially worse than reasoning-oriented models; and \textbf{(4)} anonymizing function and variable names does not affect performance (Appendix~\ref{appendix:data_contamination}).

\subsection{Precondition Prediction}
\label{subsec:further_results}





\begin{table}[t!]
    \centering
    \footnotesize
    \setlength{\tabcolsep}{10pt}
    \begin{tabular}{l | l l | l l }
        \toprule
        \textbf{Model} & \multicolumn{2}{c|}{\textbf{Domain W. ~$\uparrow$}} & \multicolumn{2}{c}{\textbf{\svcomp W. ~$\uparrow$}} \\
        & P@1  & P@3 & P@1 & P@3 \\
        \midrule
        Claude S-4.5   &  \textbf{0.87} &  \textbf{0.96 } & \textbf{0.80} & \textbf{0.83} \\
        CWM    & 0.82 & \textbf{0.96}     & 0.75 & \textbf{0.83}  \\
        Qwen3-32B    & 0.85 & 0.94 & 0.77 & \textbf{0.83}\\
        \bottomrule
    \end{tabular}
      \caption{\textbf{LLMs performance per witness formats.} We report Mean Pass@1 and Pass@3  for correct witness predictions on 40 \svcomp NT samples (10 generations each) on these short code samples. All models perform well, with higher accuracy for Domain W.
    }
    \label{tab:witness_prediction_accuracy}
\end{table}


Since the \svcomp witness format is rather complex, we propose an interpretable alternative in which models are asked to directly output a logical expression over non-deterministic variables characterizing conditions for non-termination (e.g., $x < 0 \wedge y = 0$ indicates that the program diverges only when $x$ is negative and $y$ is zero).

To assess models' ability to predict logical expressions representing non-termination (NT), we manually annotated $40$ NT programs from the \svcomp termination category. Each program is labeled with a logical expression for the NT domain.
One annotator labeled all samples, while four others annotated a disjoint subset.
Annotators were instructed to \emph{``Read the C program and specify a logical expression over the non-deterministic variables characterizing inputs leading to non-termination.''}
All annotators exhibited perfect agreement with the first annotator. We restrict to short programs with at most two non-deterministic variables, to keep annotation manageable and reliable; as a result, this base dataset contains simple, single-shot logical expressions over linear conditions.

\noindent\textbf{Formalization.} Let $P$ be a program with non-deterministic input variables $X=(x_1,\dots,x_k)$. We restrict ourselves to programs whose non-determinism is fully captured by the finite set of integer variables $X$. Thus, fixing $X=x$ determines a single execution. More general settings involving, e.g., heap non-determinism or concurrency may admit multiple executions. We write $P(x)\!\uparrow$ if the execution of $P$ under $x$ does not terminate, and define the \emph{divergence domain} as $D_P={x\in\mathbb{Z}^k:P(x)\!\uparrow}$.

\begin{definition}[Divergence Precondition]
A predicate $\varphi$ over $X$ is a \emph{sound} divergence precondition for $P$ if $\{x : \varphi(x)\} \subseteq D_P$ (every valuation satisfying $\varphi$ diverges), and a \emph{complete} divergence precondition if $\{x : \varphi(x)\} \supseteq D_P$ (every diverging valuation satisfies $\varphi$). We call $\varphi$ an \emph{exact} divergence precondition, or \emph{domain witness}, if it is both, i.e., $\{x : \varphi(x)\} = D_P$.
\end{definition}
$D_P$ corresponds to the negation of the weakest precondition of \citet{Dijkstra1976} with postcondition ``true,'' or equivalently, to the weakest precondition of a divergent Hoare triple \citep{RaadVO24}.

Our ground-truth annotations target exact divergence preconditions, i.e., the \emph{full} divergence domain $D_P$, not merely a sufficient condition for it. Since model predictions for NT domains may use different (syntactically) but equivalent logical expressions (e.g., $x<10$ vs.\ $x\leq9$), we score a prediction $\hat\varphi$ as correct iff it is logically equivalent to the ground truth $\varphi^*$, i.e., $\forall x.\ \hat\varphi(x) \leftrightarrow \varphi^*(x)$, which we check by testing the unsatisfiability of $\hat\varphi \oplus \varphi^*$ with the Z3 theorem prover~\citep{de2008z3}. A predicted $\hat\varphi$ that is sound but not complete (e.g., a strictly smaller subset of $D_P$) is therefore scored as incorrect under this exact-match criterion, even though it identifies a genuine, if partial, cause of non-termination.

Table~\ref{tab:witness_prediction_accuracy} reports mean Pass@1 and Pass@3~\citep{chen2021evaluating} over the 40 annotated samples, with 10 predictions per sample.
``Domain'' denotes our proposed format, while ``\svcomp'' uses the graph-based format. See Appendix~\ref{appendix:domain_witness_prompt} for the prompt.
Both achieve high accuracy, but Domain consistently outperforms \svcomp in Pass@1 and Pass@3, indicating stronger support for valid witness generation. It is also more interpretable.

\noindent\textbf{Extension to Non-Linear Conditions.}
To test the domain-witness format under harder, non-linear divergence conditions, we add $40$ NT programs from the \svcomp \emph{termination-nla} sub-category, involving polynomial terms (e.g., products and squares), and follow the same annotation protocol.

We evaluate Claude Sonnet-4.5, the strongest model on the original subset (Table~\ref{tab:witness_prediction_accuracy}), on this extension, using the same protocol: 10 generations per program, reporting Pass@1 and Pass@3; we restrict this extension to a single model, as our goal is a targeted stress test of the domain-witness format rather than a further cross-model comparison.

Table~\ref{tab:nla_extension} reports the results: Claude Sonnet-4.5 achieves $76.7\%$ Pass@1 and $86.5\%$ Pass@3 on the non-linear subset, compared to $87.0\%$ and $96.0\%$ on the original linear subset. All predictions were syntactically valid logical expressions; the performance drop is entirely attributable to semantic errors, primarily omitting pre-loop input guards or predicting sound but incomplete subsets of the true divergence domain. Overall, performance remains strong despite the increased difficulty, indicating reasoning beyond simple linear cases. We release the extended benchmark, annotations, and evaluation code in our public codebase.

\begin{table}[t]
    \centering
    \footnotesize
    \setlength{\tabcolsep}{10pt}
    \begin{tabular}{l | l l }
        \toprule
        \textbf{Subset} & \textbf{P@1 $\uparrow$} & \textbf{P@3 $\uparrow$} \\
        \midrule
        Linear (original, $n{=}40$)     & 0.87 & 0.96 \\
        Non-linear (extension, $n{=}40$) & 0.77 & 0.87 \\
        \bottomrule
    \end{tabular}
    \caption{\textbf{Claude Sonnet-4.5 on linear vs.\ non-linear domain witnesses.} Mean Pass@1 and Pass@3 (10 generations per program) on the original linear subset and the non-linear (\svcomp \texttt{termination-nla}).}
    \label{tab:nla_extension}
\end{table}

\paragraph{Examples of model witness outputs.} We find that certain NT samples yield a valid domain witness format but not a valid graph-based format, and vice versa. This suggests that the two formats are complementary: errors in one format do not necessarily occur in the other.
See Appendix~\ref{appendix:model_witness_responses} 
for examples of both formats, including one with the model's reasoning process prior to its final answer.




\section{Related Work}
\label{sec:related_work}

\noindent\textbf{LLMs vs. classic reasoners.} \citet{kambhampati2024can} examines the capacity of LLMs for reasoning and planning, showing that they underperform classical planners on benchmarks from the International Planning Competition~\citep{o1PlanBench}. They argue that while LLMs may complement traditional planners through their language capabilities, they fall short in both accuracy and efficiency~\citep{LLM-modulo}. Similarly, \citet{Hazra-SAT} report that reasoning-oriented models yield improvements on SAT tasks but still do not reach the performance of classical solvers.
In the context of theorem proving, prior work mostly integrates LLMs with symbolic reasoning systems such as Lean, SMT solvers, and first-order logic provers-where LLMs generate candidate proofs that are verified by symbolic tools, improving performance and reliability \citep{sultan2025towards, trinh2024solving, mirzadeh2024gsm, regin2024combining, szeider2024mcp, olausson2023linc}. In contrast, we evaluate the standalone ability of LLMs to reason about program termination, noting that practical systems should pair them with symbolic proof checkers.

\noindent\textbf{Termination analysis using LLMs.} Most prior work has focused on predicting program inputs and outputs \citep{CruxEval, xu2025cruxeval}. 
Recent studies show that jointly training models to predict code and execution traces improves code generation \citep{armengol2025cannot}, and that learning code world models further extends this capability \citep{carbonneaux2025cwm}. Termination prediction goes beyond execution, requiring reasoning about the existence of non-terminating behaviors.
Another line of work uses LLMs to synthesize termination arguments, such as ranking functions \citep{MSFTProgVer}, but, to our knowledge, does not address non-termination witnesses. \citet{DBLP:conf/sigsoft/AlonD22} apply GNNs to termination reasoning, using timeouts as a heuristic proxy for non-termination. More recently, \citet{carbonneaux2025cwm} evaluate CWM and Qwen3 on a simplified subset of translated Python functions from \svcomp and TPDB,~\footnote{\url{https://github.com/TermCOMP/TPDB}} showing strong performance.

\noindent\textbf{Termination analysis tools.} In \svcomp~2025, the leading systems are PROTON~\citep{metta2024proton} and UAutomizer~\citep{heizmann2023ultimate}. PROTON uses CBMC~\citep{kroening2014cbmc} to soundly detect NT via recurrent loop states and generate witnesses, and falls back on high-confidence termination checks from VeriFuzz~\citep{metta2023verifuzz}. UAutomizer encodes programs as automata and applies SMT-based refinement to eliminate infeasible paths.
\svcomp tools typically target small, self-contained programs. In contrast, \citet{RaadVO24} study repository-level termination over millions of lines of code using recurrent sets~\citep{Gupta08}, bi-abduction~\citep{Calcagno11}, and under-approximation~\citep{OHearn20}.



\section{Discussion}
\label{sec:discussion}


\noindent\textbf{Decidable vs. Undecidable problems.} As discussed in Section~\ref{sec:related_work}, LLMs underperform classical solvers on decidable tasks such as SAT and propositional planning. Interestingly, our results suggest that frontier LLMs can achieve competitive empirical performance with specialized tools on practical termination benchmarks, a task that is theoretically more challenging due to its undecidability.
This discrepancy may be partly attributable to differences in benchmark construction: SAT and planning competition benchmarks are largely synthetically generated, whereas \svcomp has evolved through community contributions. As a result, \svcomp more closely reflects naturally occurring verification problems rather than being explicitly designed to be adversarial.
Additionally, undecidable problems depend on
heuristic reasoning rather than exhaustive search, making them potentially a more natural fit for LLMs.

\noindent\textbf{Beyond \svcomp.} 
\svcomp remains the most comprehensive and challenging benchmark for termination analysis.
Prior work focused on less-challenging benchmarks, e.g., on Python programs translated from TPDB~\citep{carbonneaux2025cwm}. 
However, we believe that making LLMs useful for program termination requires extending to even more challenging settings, and that the ultimate assessment of the practical utility of LLMs in termination analysis should derive from deployment on real-world codebases (e.g., within continuous integration). In practical deployments, automatic termination reasoning must handle a continual stream of previously unseen problems and meet the accuracy requirements of real-world engineering.

\section{Conclusions and Future Work}
\label{sec:conclusions_and_future_work}

We studied whether frontier LLMs can reason about program termination competitively with state-of-the-art verification tools on the \svcomp benchmark. We empirically show strong performance of LLMs -- GPT-5 and Claude Sonnet-4.5 (with TTS) achieve performance close to the top-ranked verifier PROTON, while CWM performs close to UAutomizer. Across all models, performance degrades as code length increases, and witness generation remains a notable challenge.

To address this gap, we introduced a divergence-precondition formulation that characterizes non-termination conditions as logical constraints. Notably, LLMs predict these preconditions through prompting, whereas classical verifiers designed for binary termination decisions do not support this related task out of the box, highlighting the flexibility of LLM-based program analysis.

More broadly, our work highlights program termination as a useful testbed for understanding the strengths and limitations of frontier LLM reasoning, particularly the emerging gap between semantic recognition and formally verifiable reasoning.

In the future, we plan to construct challenging real-world termination benchmarks, explore neuro-symbolic approaches that combine LLMs with symbolic verifiers, and extend this line of research to additional undecidable reasoning problems.





\section*{Limitations}

\begin{itemize}
    \item \textbf{Programming Language.} Our evaluation focuses on \svcomp, which consists exclusively of $C$ programs. Notably, prior work suggests that LLM-based termination reasoning can transfer across languages, e.g., to Python programs translated from TPDB~\cite{carbonneaux2025cwm}.

    \item \textbf{Proprietary LLM (OpenAI GPT-5).} 
    GPT-5 achieves the best performance across models. While this model delivers strong performance, its architecture, training data, and training methodology are proprietary.  
    \item \textbf{LLMs are sensitive to prompt phrasing.} LLMs are sometimes sensitive to small changes to the prompts. We acknowledge that results may vary with different prompts. We note that we use a simple standardized prompt without benchmark-specific tuning to avoid overfitting. Despite this minimal setup, models achieve competitive performance.

    \item \textbf{Lab versus competition conditions.} The LLM results reported on in this work were not from official entry to the \svcomp competition, and exact competition conditions cannot be guaranteed, and this could affect comparison to the scores of \svcomp entrants. We expect that our LLM results are conservative, as we use one witness validator rather than many. This work was completed after the deadline for \svcomp 2026; perhaps LLM entries could participate in \svcomp 2027. 

    \item \textbf{Compute comparability.} LLM inference (API calls, or GPU inference for open-weight models) and the CPU-bound \svcomp verification tools are evaluated under different computational conditions and are not directly cost-comparable; \svcomp-TTS additionally requires multiple LLM generations per instance. We report token consumption per model as a reproducible, hardware-independent proxy for LLM inference cost (Appendix~\ref{appendix:compute_cost}).

    \item \textbf{Robustness evaluation.} We leave a more systematic robustness evaluation under semantics-preserving transformations, adversarial rewrites, and procedurally generated benchmarks to future work. Consequently, our findings should primarily be interpreted as a behavioral characterization of current frontier LLMs on \svcomp-style termination tasks rather than definitive evidence of general semantic reasoning capabilities.

\end{itemize}

\label{sec:limitations}

\section*{Ethical Considerations}

\noindent\textbf{AI Assistants.}  
We used AI assistants for coding support, writing, and rephrasing. All AI-generated outputs were carefully reviewed and edited to ensure alignment with our design goals and to preserve the original intent of our work.

\section*{Acknowledgments}
We thank Ori Yoran for his valuable feedback, as well as the whole CodeGen team at Meta FAIR for helpful discussions. We thank Marco Selvi and Kunhao Zheng for the initial integration of the UAutomizer witness checker.


\bibliography{custom}

@book{Dijkstra1976,
  author    = {Edsger W. Dijkstra},
  title     = {A Discipline of Programming},
  publisher = {Prentice Hall},
  year      = {1976},
  address   = {Englewood Cliffs, NJ}
}

@article{OHearn20,
  author       = {Peter W. O'Hearn},
  title        = {Incorrectness logic},
  journal      = {Proc. {ACM} Program. Lang.},
  volume       = {4},
  number       = {{POPL}},
  pages        = {10:1--10:32},
  year         = {2020},
  url          = {https://doi.org/10.1145/3371078},
  doi          = {10.1145/3371078},
  bibsource    = {dblp computer science bibliography, https://dblp.org}
}

@article{Calcagno11,
  author       = {Cristiano Calcagno and
                  Dino Distefano and
                  Peter W. O'Hearn and
                  Hongseok Yang},
  title        = {Compositional Shape Analysis by Means of Bi-Abduction},
  journal      = {J. {ACM}},
  volume       = {58},
  number       = {6},
  pages        = {26:1--26:66},
  year         = {2011},
  url          = {https://doi.org/10.1145/2049697.2049700},
  doi          = {10.1145/2049697.2049700},
  bibsource    = {dblp computer science bibliography, https://dblp.org}
}

@inproceedings{Gupta08,
  author       = {Ashutosh Gupta and
                  Thomas A. Henzinger and
                  Rupak Majumdar and
                  Andrey Rybalchenko and
                  Ru{-}Gang Xu},
  editor       = {George C. Necula and
                  Philip Wadler},
  title        = {Proving non-termination},
  booktitle    = {Proceedings of the 35th {ACM} {SIGPLAN-SIGACT} Symposium on Principles
                  of Programming Languages, {POPL} 2008, San Francisco, California,
                  USA, January 7-12, 2008},
  pages        = {147--158},
  publisher    = {{ACM}},
  year         = {2008},
  url          = {https://doi.org/10.1145/1328438.1328459},
  doi          = {10.1145/1328438.1328459},
  bibsource    = {dblp computer science bibliography, https://dblp.org}
}

@article{RaadVO24,
  author       = {Azalea Raad and
                  Julien Vanegue and
                  Peter W. O'Hearn},
  title        = {Non-termination Proving at Scale},
  journal      = {Proc. {ACM} Program. Lang.},
  volume       = {8},
  number       = {{OOPSLA2}},
  pages        = {246--274},
  year         = {2024},
  url          = {https://doi.org/10.1145/3689720},
  doi          = {10.1145/3689720},
  bibsource    = {dblp computer science bibliography, https://dblp.org}
}

@article{o1PlanBench,
  author       = {Karthik Valmeekam and
                  Kaya Stechly and
                  Subbarao Kambhampati},
  title        = {LLMs Still Can't Plan; Can LRMs? {A} Preliminary Evaluation of
                  OpenAI's o1 on PlanBench},
  journal      = {CoRR},
  volume       = {abs/2409.13373},
  year         = {2024},
  url          = {https://doi.org/10.48550/arXiv.2409.13373},
  doi          = {10.48550/ARXIV.2409.13373},
  eprinttype    = {arXiv},
  eprint       = {2409.13373},
  bibsource    = {dblp computer science bibliography, https://dblp.org}
}

@article{Hazra-SAT,
  author       = {Rishi Hazra and
                  Gabriele Venturato and
                  Pedro Zuidberg Dos Martires and
                  Luc De Raedt},
  title        = {Have Large Language Models Learned to Reason? {A} Characterization
                  via 3-SAT Phase Transition},
  journal      = {CoRR},
  volume       = {abs/2504.03930},
  year         = {2025},
  url          = {https://doi.org/10.48550/arXiv.2504.03930},
  doi          = {10.48550/ARXIV.2504.03930},
  eprinttype    = {arXiv},
  eprint       = {2504.03930},
  bibsource    = {dblp computer science bibliography, https://dblp.org}
}

@inproceedings{LLM-modulo,
  title={Position: LLMs can’t plan, but can help planning in LLM-modulo frameworks},
  author={Kambhampati, Subbarao and Valmeekam, Karthik and Guan, Lin and Verma, Mudit and Stechly, Kaya and Bhambri, Siddhant and Saldyt, Lucas Paul and Murthy, Anil B},
  booktitle={Forty-first International Conference on Machine Learning},
  year={2024}
}

@inproceedings{MSFTProgVer,
  author       = {Adharsh Kamath and
                  Nausheen Mohammed and
                  Aditya Senthilnathan and
                  Saikat Chakraborty and
                  Pantazis Deligiannis and
                  Shuvendu K. Lahiri and
                  Akash Lal and
                  Aseem Rastogi and
                  Subhajit Roy and
                  Rahul Sharma},
  editor       = {Nina Narodytska and
                  Philipp R{\"{u}}mmer},
  title        = {Leveraging LLMs for Program Verification},
  booktitle    = {Formal Methods in Computer-Aided Design, {FMCAD} 2024, Prague, Czech
                  Republic, October 15-18, 2024},
  pages        = {107--118},
  publisher    = {{IEEE}},
  year         = {2024},
  url          = {https://doi.org/10.34727/2024/isbn.978-3-85448-065-5\_16},
  doi          = {10.34727/2024/ISBN.978-3-85448-065-5\_16},
  bibsource    = {dblp computer science bibliography, https://dblp.org}
}

@inproceedings{beyer2012competition,
  title={Competition on Software Verification: (SV-COMP)},
  author={Beyer, Dirk},
  booktitle={International Conference on Tools and Algorithms for the Construction and Analysis of Systems},
  pages={504--524},
  year={2012},
  organization={Springer}
}

@inproceedings{CruxEval,
  author       = {Alex Gu and
                  Baptiste Rozi{\`{e}}re and
                  Hugh James Leather and
                  Armando Solar{-}Lezama and
                  Gabriel Synnaeve and
                  Sida Wang},
  title        = {CRUXEval: {A} Benchmark for Code Reasoning, Understanding and Execution},
  booktitle    = {Forty-first International Conference on Machine Learning, {ICML} 2024,
                  Vienna, Austria, July 21-27, 2024},
  publisher    = {OpenReview.net},
  year         = {2024},
  url          = {https://openreview.net/forum?id=Ffpg52swvg},
  bibsource    = {dblp computer science bibliography, https://dblp.org}
}

@inproceedings{DBLP:conf/sigsoft/AlonD22,
  author       = {Yoav Alon and
                  Cristina David},
  editor       = {Abhik Roychoudhury and
                  Cristian Cadar and
                  Miryung Kim},
  title        = {Using graph neural networks for program termination},
  booktitle    = {Proceedings of the 30th {ACM} Joint European Software Engineering
                  Conference and Symposium on the Foundations of Software Engineering,
                  {ESEC/FSE} 2022, Singapore, Singapore, November 14-18, 2022},
  pages        = {910--921},
  publisher    = {{ACM}},
  year         = {2022},
  url          = {https://doi.org/10.1145/3540250.3549095},
  doi          = {10.1145/3540250.3549095},
  bibsource    = {dblp computer science bibliography, https://dblp.org}
}

@inproceedings{de2008z3,
  title={Z3: An efficient SMT solver},
  author={De Moura, Leonardo and Bj{\o}rner, Nikolaj},
  booktitle={International conference on Tools and Algorithms for the Construction and Analysis of Systems},
  pages={337--340},
  year={2008},
  organization={Springer}
}

@article{carbonneaux2025cwm,
  title={CWM: An Open-Weights LLM for Research on Code Generation with World Models},
  author={Copet, Jade and Carbonneaux, Quentin and Cohen, Gal and Gehring, Jonas and Kahn, Jacob and Kossen, Jannik and Kreuk, Felix and McMilin, Emily and Meyer, Michel and Wei, Yuxiang and others},
  journal={arXiv preprint arXiv:2510.02387},
  year={2025}
}

@article{yang2025qwen3,
  title={Qwen3 technical report},
  author={Yang, An and Li, Anfeng and Yang, Baosong and Zhang, Beichen and Hui, Binyuan and Zheng, Bo and Yu, Bowen and Gao, Chang and Huang, Chengen and Lv, Chenxu and others},
  journal={arXiv preprint arXiv:2505.09388},
  year={2025}
}

@misc{claude4.5,
    title = {Claude Sonnet 4.5},
    url = {https://www.anthropic.com/news/claude-sonnet-4-5},
    author = {Anthropic},
    month = {September},
    year = {2025}
}

@misc{gpt-5,
    title = {GPT-5},
    url = {https://openai.com/index/introducing-gpt-5/},
    author = {OpenAI},
    month = {August},
    year = {2025}
}

@article{hurst2024gpt,
  title={Gpt-4o system card},
  author={Hurst, Aaron and Lerer, Adam and Goucher, Adam P and Perelman, Adam and Ramesh, Aditya and Clark, Aidan and Ostrow, AJ and Welihinda, Akila and Hayes, Alan and Radford, Alec and others},
  journal={arXiv preprint arXiv:2410.21276},
  year={2024}
}

@inproceedings{metta2024proton,
  title={PROTON: probes for termination or not (competition contribution)},
  author={Metta, Ravindra and Karmarkar, Hrishikesh and Madhukar, Kumar and Venkatesh, R and Chakraborty, Supratik},
  booktitle={International Conference on Tools and Algorithms for the Construction and Analysis of Systems},
  pages={393--398},
  year={2024},
  organization={Springer}
}

@inproceedings{heizmann2023ultimate,
  title={Ultimate Automizer and the CommuHash Normal Form: (Competition Contribution)},
  author={Heizmann, Matthias and Barth, Max and Dietsch, Daniel and Fichtner, Leonard and Hoenicke, Jochen and Klumpp, Dominik and Naouar, Mehdi and Schindler, Tanja and Sch{\"u}ssele, Frank and Podelski, Andreas},
  booktitle={International Conference on Tools and Algorithms for the Construction and Analysis of Systems},
  pages={577--581},
  year={2023},
  organization={Springer}
}

@inproceedings{kroening2014cbmc,
  title={CBMC--C Bounded Model Checker: (Competition Contribution)},
  author={Kroening, Daniel and Tautschnig, Michael},
  booktitle={International Conference on Tools and Algorithms for the Construction and Analysis of Systems},
  pages={389--391},
  year={2014},
  organization={Springer}
}

@inproceedings{metta2023verifuzz,
  title={VeriFuzz 1.4: Checking for (non-) termination (competition contribution)},
  author={Metta, Ravindra and Yeduru, Prasanth and Karmarkar, Hrishikesh and Medicherla, Raveendra Kumar},
  booktitle={International Conference on Tools and Algorithms for the Construction and Analysis of Systems},
  pages={594--599},
  year={2023},
  organization={Springer}
}

@article{turing1936computable,
  title={On computable numbers, with an application to the Entscheidungsproblem},
  author={Turing, Alan Mathison and others},
  journal={J. of Math},
  volume={58},
  number={345-363},
  pages={5},
  year={1936},
  publisher={Wiley Online Library}
}

@article{avizienis2004basic,
  title={Basic concepts and taxonomy of dependable and secure computing},
  author={Avizienis, Algirdas and Laprie, J-C and Randell, Brian and Landwehr, Carl},
  journal={IEEE transactions on dependable and secure computing},
  volume={1},
  number={1},
  pages={11--33},
  year={2004},
  publisher={IEEE}
}

@article{zheng2023survey,
  title={A survey of large language models for code: Evolution, benchmarking, and future trends},
  author={Zheng, Zibin and Ning, Kaiwen and Wang, Yanlin and Zhang, Jingwen and Zheng, Dewu and Ye, Mingxi and Chen, Jiachi},
  journal={arXiv preprint arXiv:2311.10372},
  year={2023}
}

@article{zhao2023survey,
  title={A survey of large language models},
  author={Zhao, Wayne Xin and Zhou, Kun and Li, Junyi and Tang, Tianyi and Wang, Xiaolei and Hou, Yupeng and Min, Yingqian and Zhang, Beichen and Zhang, Junjie and Dong, Zican and others},
  journal={arXiv preprint arXiv:2303.18223},
  volume={1},
  number={2},
  year={2023}
}

@article{liu2024deepseek,
  title={Deepseek-v3 technical report},
  author={Liu, Aixin and Feng, Bei and Xue, Bing and Wang, Bingxuan and Wu, Bochao and Lu, Chengda and Zhao, Chenggang and Deng, Chengqi and Zhang, Chenyu and Ruan, Chong and others},
  journal={arXiv preprint arXiv:2412.19437},
  year={2024}
}

@article{jiang2024survey,
  title={A survey on large language models for code generation},
  author={Jiang, Juyong and Wang, Fan and Shen, Jiasi and Kim, Sungju and Kim, Sunghun},
  journal={arXiv preprint arXiv:2406.00515},
  year={2024}
}

@article{emrich2023aprove,
  title={AProVE: Modular Termination Analysis of Memory-Manipulating C Programs},
  author={Emrich, Frank and Hensel, Jera and Giesl, J{\"u}rgen},
  journal={arXiv preprint arXiv:2302.02382},
  year={2023}
}

@article{kambhampati2024can,
  title={Can large language models reason and plan?},
  author={Kambhampati, Subbarao},
  journal={Annals of the New York Academy of Sciences},
  volume={1534},
  number={1},
  pages={15--18},
  year={2024},
  publisher={Wiley Online Library}
}

@article{sultan2025towards,
  title={Towards Reliable Proof Generation with LLMs: A Neuro-Symbolic Approach},
  author={Sultan, Oren and Stern, Eitan and Shahaf, Dafna},
  journal={arXiv preprint arXiv:2505.14479},
  year={2025}
}

@article{szeider2024mcp,
  title={MCP-solver: Integrating language models with constraint programming systems},
  author={Szeider, Stefan},
  journal={arXiv preprint arXiv:2501.00539},
  year={2024}
}

@article{regin2024combining,
  title={Combining constraint programming reasoning with large language model predictions},
  author={R{\'e}gin, Florian and De Maria, Elisabetta and Bonlarron, Alexandre},
  journal={arXiv preprint arXiv:2407.13490},
  year={2024}
}

@article{mirzadeh2024gsm,
  title={Gsm-symbolic: Understanding the limitations of mathematical reasoning in large language models},
  author={Mirzadeh, Iman and Alizadeh, Keivan and Shahrokhi, Hooman and Tuzel, Oncel and Bengio, Samy and Farajtabar, Mehrdad},
  journal={arXiv preprint arXiv:2410.05229},
  year={2024}
}

@article{olausson2023linc,
  title={LINC: A neurosymbolic approach for logical reasoning by combining language models with first-order logic provers},
  author={Olausson, Theo X and Gu, Alex and Lipkin, Benjamin and Zhang, Cedegao E and Solar-Lezama, Armando and Tenenbaum, Joshua B and Levy, Roger},
  journal={arXiv preprint arXiv:2310.15164},
  year={2023}
}

@article{trinh2024solving,
  title={Solving olympiad geometry without human demonstrations},
  author={Trinh, Trieu H and Wu, Yuhuai and Le, Quoc V and He, He and Luong, Thang},
  journal={Nature},
  volume={625},
  number={7995},
  pages={476--482},
  year={2024},
  publisher={Nature Publishing Group UK London}
}

@article{chen2021evaluating,
  title={Evaluating large language models trained on code},
  author={Chen, Mark},
  journal={arXiv preprint arXiv:2107.03374},
  year={2021}
}

@article{armengol2025cannot,
  title={What I cannot execute, I do not understand: Training and Evaluating LLMs on Program Execution Traces},
  author={Armengol-Estap{\'e}, Jordi and Carbonneaux, Quentin and Zhang, Tianjun and Markosyan, Aram H and Seeker, Volker and Cummins, Chris and Kambadur, Melanie and O'Boyle, Michael FP and Wang, Sida and Synnaeve, Gabriel and others},
  journal={arXiv preprint arXiv:2503.05703},
  year={2025}
}

@inproceedings{xu2025cruxeval,
  title={Cruxeval-x: A benchmark for multilingual code reasoning, understanding and execution},
  author={Xu, Ruiyang and others},
  booktitle={Proceedings of the 63rd Annual Meeting of the Association for Computational Linguistics (Volume 1: Long Papers)},
  pages={23762--23779},
  year={2025}
}

@article{rice1953classes,
  title={Classes of recursively enumerable sets and their decision problems},
  author={Rice, Henry Gordon},
  journal={Transactions of the American Mathematical society},
  volume={74},
  number={2},
  pages={358--366},
  year={1953},
  publisher={JSTOR}
}

@inproceedings{beyer2025improvements,
  title={Improvements in software verification and witness validation: SV-COMP 2025},
  author={Beyer, Dirk and Strej{\v{c}}ek, Jan},
  booktitle={International Conference on Tools and Algorithms for the Construction and Analysis of Systems},
  pages={151--186},
  year={2025},
  organization={Springer}
}

\appendix
\label{appendix}

\section{Reproducibility}

The \svcomp dataset we used is available at: \url{https://gitlab.com/sosy-lab/sv-comp/bench-defs}.  
To obtain the benchmarks, clone the repository.
Then, checkout the \emph{svcomp25} tag and initialize submodules (including \emph{sv-benchmarks}).  
Repository versions used in the competition are listed in Table 4 of the report: \url{https://doi.org/10.1007/978-3-031-90660-2_9}.
Benchmarks are organized by property (e.g., \emph{Termination}), with sub-categories.  
We followed the official instructions and excluded $11$ invalid tasks ($9$ from ``MainControlFlow'', $2$ from ``Other''), resulting in $2{,}328$ samples.
\label{appendix:reproducibility}

\section{Dataset Statistics}

\noindent\textbf{\svcomp Termination Sub-categories.} 
See subcategories summarization in Table~\ref{tab:termination_category_summary}.

\begin{figure*}[t]
\includegraphics[width=0.5\textwidth]{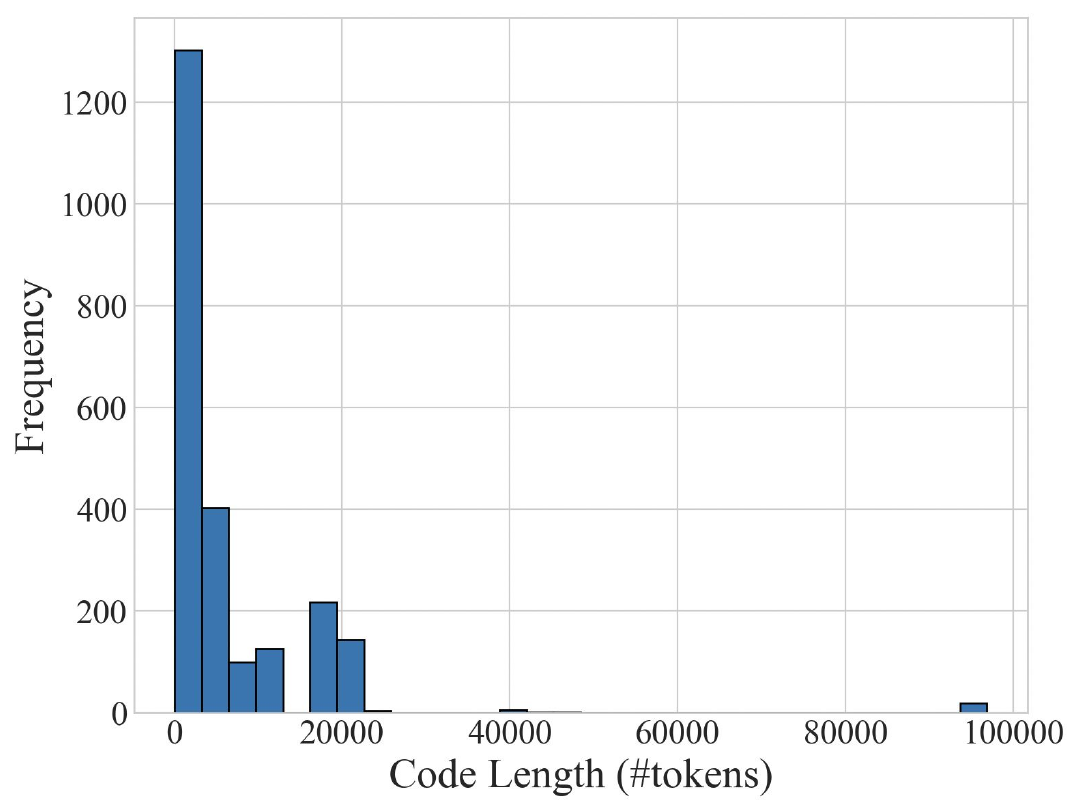}
\centering
\caption{
Histogram of input code length in \svcomp 2025. \textbf{X axis:} code length in tokens. \textbf{Y axis:} frequency in the data.
Most code samples have $20K$ tokens ($\mu=6,080$, $\sigma=10,898$).
}
\label{fig:code_length_hist}
\end{figure*}

\noindent\textbf{Length distribution.} The dataset exhibits significant variability in code length: token counts range from $45$ to $96{,}827$, with a mean of $\mu = 6{,}080$ and a standard deviation of $\sigma = 10{,}898$. See Figure~\ref{fig:code_length_hist} for the code length histogram.

\label{appendix:dataset_stats}

\section{\svcomp Witness}

According to the \svcomp specification, when a non-termination is predicted, a \texttt{witness} must be provided to encode the proof as a graph structure. In this graph, nodes represent distinct program states, each assigned a unique \texttt{id}. Nodes may also include special attributes such as \texttt{entry} (marking the entry point) or \texttt{cyclehead} (indicating the head of a cycle). Edges denote transitions between states, corresponding to lines of code. Each edge includes its own \texttt{id}, the relevant source code line, the \texttt{source} and \texttt{target} node IDs, the actual \texttt{sourcecode} line, and any special properties (e.g., \texttt{enterLoopHead}, \texttt{control}, \texttt{assumption}).

\begin{figure*}[t]
    \includegraphics[width=1.0\textwidth]{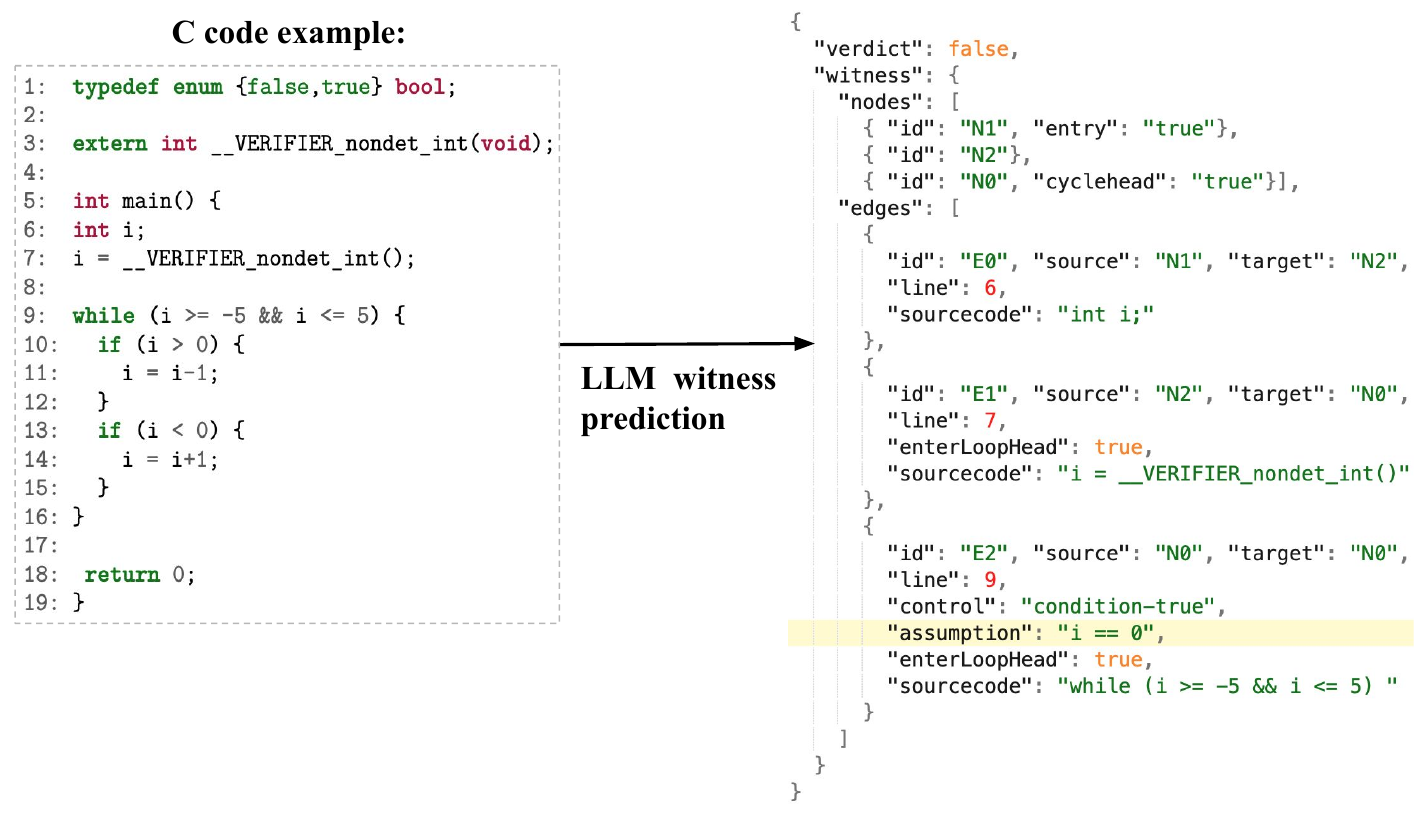}
    \centering
    \caption{Example of an LLM-generated witness automaton (graph) in json format.}
    \label{appendix:llm_witness_json}
\end{figure*}

To generate this witness, we prompt the LLM to produce a JSON object that encodes the graph. This JSON is subsequently converted to a GraphML file and validated using a dedicated verifier tool. For a visual illustration, refer to Figure~\ref{appendix:llm_witness_json}.
\label{appendix:svcomp_witness}

\section{\svcomp Prompt}

See Figure~\ref{fig:termination_prompt} for a detailed description of the prompt provided to the LLM, including the termination analysis task instructions and the requirements for witness proofs in \svcomp format for non-termination predictions. See Figure~\ref{fig:termination_few_shot} for 
few-shot examples given to the LLM. These include one example demonstrating non-termination and another demonstrating termination.



\label{appendix:svcomp_prompt}

\section{UAutomizer Witness Validator}

Ultimate Automizer (UAutomizer) is a powerful verification tool capable of both producing and validating witnesses for program verification tasks. As a witness validator, UAutomizer checks whether a given witness automaton (encoded in GraphML) correctly describes a feasible error path (violation witness) or a proof of correctness (correctness witness) for a program and its specification.

\paragraph{Validation workflow.}
UAutomizer requires the program source file (e.g., \emph{test.c}), the specification file (in our  task, \emph{../properties/termination.prp}), the witness automaton in GraphML format (e.g., \emph{witness.graphml}), and the architecture ($32$ or $64$ bit).

\paragraph{Command-line invocation.}
\begin{verbatim}
./Ultimate.py \
  --architecture 32bit \
  --spec PropertyUnreachCall.prp \
  --file test.c \
  --validate witness.graphml
\end{verbatim}

\paragraph{Validation mechanism.} UAutomizer parses the witness automaton (GraphML) and interprets its nodes and edges according to the witness specification. It simulates program execution, guided by the automaton. It checks whether its error path corresponds to a feasible program execution that violates the specification.

\paragraph{Result interpretation.} If UAutomizer finds a matching error path, it outputs \texttt{FALSE}, meaning the witness is validated and the program violates the specification as described. If no such path exists, or the automaton does not match the program, it outputs \texttt{TRUE}, meaning the witness is rejected.

\paragraph{Interpreting the witness automaton.}
UAutomizer expects the witness automaton to be encoded in GraphML, with nodes representing control states and edges representing transitions. It reads \texttt{<data>} elements attached to nodes and edges, which encode information such as entry/sink/violation states, invariants, assumptions, source code locations, and thread information. The automaton must include specification and architecture information, allowing UAutomizer to match the witness to the correct verification context. 
See \url{https://github.com/sosy-lab/sv-witnesses/blob/main/README-GraphML.md} for reference.

\label{appendix:UAutomizer}

\section{Code Length Error Analysis}

One of the key questions in our error analysis is whether longer code snippets are associated with a higher rate of mistakes across models. Table~\ref{tab:mean_num_tokens_score_type} reports the mean code length for each prediction outcome. Our results show that incorrect predictions (false negatives and false positives) are indeed linked to longer inputs, while correct predictions (true positives and true negatives) are more common for shorter code snippets. Additionally, unknowns and unconfirmed witnesses are also associated with longer code, suggesting that models are more likely to abstain from answering when confronted with complex or lengthy programs.

\label{appendix:errors_analysis}

\section{Sub-categories Error Rates Analysis}
We analyze GPT-5, the best-performing model. 
Errors are concentrated in a few subcategories. While \texttt{openssl-simplified} (100\%) and \texttt{ldv-sets} (60\%) exhibit the highest failure rates, they represent very small portions of the dataset. In contrast, multi-threaded programs (e.g., \texttt{seq-mthreaded}, $\sim$6\%) remain a primary weakness, with a high failure rate (15\%) dominated by false positives.
Additional challenges appear in \texttt{systemc} ($\sim$2.5\%, 7.9\%, mainly parsing failures) and heap-based programs (e.g., linked lists, $\sim$0.7\%, 5.9\%), an association we take as suggestive, not explanatory, of difficulties with input structure and memory reasoning. In contrast, \texttt{product-lines}, the largest subcategory ($\sim$26\%), has a low failure rate (1.3\%), and simpler control-flow categories (loops, arrays, recursion) show near-zero failure rates.
\footnote{See the SV-Benchmarks repository for subcategory details: \url{https://gitlab.com/sosy-lab/benchmarking/sv-benchmarks/-/tree/main/c/}.}
\label{appendix:subcat_errors}

\section{Data Contamination Analysis}

We provide several pieces of evidence suggesting contamination
is unlikely:

\paragraph{Capability gap.}
Non-reasoning models (e.g., GPT-4o) perform substantially worse than
reasoning models (Figure~\ref{fig:svcomp_main_results}), and smaller models
frequently predict \texttt{Unknown} (Table~\ref{tab:errors_analysis_unk}),
indicating reliance on reasoning rather than lookup.

\paragraph{Difficulty sensitivity.}
Performance degrades with program complexity
(Figure~\ref{fig:code_length_bins}), inconsistent with memorization.

\paragraph{Witness validity.}
Even correct predictions often produce invalid witnesses
(Table~\ref{tab:errors_analysis_witness}), suggesting incomplete
reasoning rather than retrieval.

\paragraph{Out-of-distribution signal.}
Models recover non-termination domains in a novel format not available
online (Section~\ref{subsec:further_results}).

\paragraph{Cross-benchmark robustness.}
On Halteval-Prelim~\citep{carbonneaux2025cwm}, reasoning models outperform non-reasoning models, with no degradation from identifier anonymization.
\label{appendix:data_contamination}

\section{Compute Cost Analysis}

We measure prompt and output token consumption for each evaluated model across the full \svcomp benchmark ($2{,}328$ instances $\times$ $20$ generations $=$ $46{,}560$ predictions per model). Prompt tokens correspond to the complete model input (system prompt, task instructions, and input $C$ program), measured using each model's native tokenizer. Output tokens include all generated tokens available in the inference logs, i.e., reasoning traces (when exposed by the model) plus the final answer.

\begin{table*}[h]
    \centering
    
    \setlength{\tabcolsep}{5pt}
    \begin{tabular}{l l | c c | c c}
        \toprule
        \textbf{Model} & \textbf{Tokenizer} & \multicolumn{2}{c|}{\textbf{Prompt}} & \multicolumn{2}{c}{\textbf{Output}} \\
        & & Mean & Med. & Mean & Med. \\
        \midrule
        Claude Sonnet-4.5 & Anthropic$^*$ & 10{,}497 & 2{,}640 & 1{,}373 & 1{,}413 \\
        GPT-5             & o200k\_base   & 9{,}324  & 2{,}429 & 130$^{**}$   & 9$^{**}$ \\
        GPT-4o            & o200k\_base   & 9{,}324  & 2{,}429 & 200      & 169 \\
        CWM               & Llama-3       & 9{,}233  & 2{,}373 & 1{,}474  & 1{,}886 \\
        Qwen3-30B         & Qwen          & 11{,}009 & 2{,}467 & 1{,}125  & 884 \\
        \bottomrule
    \end{tabular}
    \caption{\textbf{Token consumption per model, main \svcomp evaluation.} Prompt and output token counts (mean, median) per generation, aggregated over the full benchmark.
    $^*$Anthropic does not publicly release the Claude 4.x tokenizer; Anthropic token counts are estimated using LiteLLM's tokenizer approximation.
    $^{**}$GPT-5 (medium reasoning effort) does not expose its hidden reasoning tokens through the API, so its reported output length undercounts total generated tokens.}
    \label{tab:compute_cost_main}
\end{table*}

Average prompt lengths range from $9.2$K-$11.0$K tokens across tokenizers, while outputs vary substantially: Claude Sonnet-4.5, CWM, and Qwen3-30B average $1.37$K, $1.47$K, and $1.12$K tokens per instance, respectively, versus $200$ for the non-reasoning GPT-4o. We report token consumption as a reproducible, hardware-independent measure of inference cost, though it is not directly comparable to the CPU-bound \svcomp tools (Table~\ref{tab:compute_cost_main}).

\paragraph{Witness validator scope.} As stated in Appendix~\ref{appendix:main_results} and in Limitations, we validate witnesses using only the UAutomizer validator, whereas the official \svcomp competition accepts a witness if validated by \emph{any} of several validators. Our reported LLM scores are therefore a conservative (stricter) lower bound relative to what an official \svcomp entry would likely achieve.

\label{appendix:compute_cost}

\section{Domain Witness Prompt}

See Figure~\ref{fig:domain_witness_prompt} for a detailed description of the prompt provided to the LLM for the domain prediction witness settings.
\label{appendix:domain_witness_prompt}

\section{Main Results}

See Table~\ref{tab:model_performance_detailed} for the main results of LLM evaluation in \svcomp 2025. The table reports mean scores and standard deviations across 100 bootstrap runs. Notably, the standard deviation for F1 scores across models is consistently low, which indicates stable classification performance, particularly for top-performing LLMs. In contrast, \svcomp score standard deviations span a much wider range, from 47 up to 750. While leading models demonstrate robust and consistent \svcomp results with lower variability, mid-tier and baseline models exhibit greater fluctuations and less reliable verification performance.


Notice, in \svcomp 2025, multiple witness validators are used, and a prediction is accepted if any of them validates it. We consider the UAutomizer validator only, uautomizer, making the LLM's evaluation \emph{stricter} and possibly yielding a lower scores compared to the official competition score.

\label{appendix:main_results}

\section{Model Witness Responses Examples}

In Section~\ref{subsec:further_results}, we sampled 40 short code snippet examples from the \texttt{svcomp} dataset and evaluated several LLMs, generating 10 predictions per sample.
Our results indicate that LLMs demonstrate strong performance in both the graph-based witness format and the domain witness format. This is expected, as this subset of code examples consists of relatively easy cases. Notably, LLMs consistently achieve higher accuracy in the domain witness format.
Interestingly, we observed instances where predictions were correct in the domain witness format but incorrect in the automaton (graph) witness format. These findings suggest that the domain witness and automaton (graph) witness formats are complementary; a mistake in one format does not necessarily imply a mistake in the other.
We present examples of CWM predictions. Both domain and graph-based witness predictions are correct (see Figure~\ref{fig:cmw_domain_correct_graph_correct}). Domain witness prediction is incorrect, but graph-based prediction is correct (see Figure~\ref{fig:cwm_domain_incorrect_graph_correct}). Domain witness prediction is correct, but graph-based prediction is incorrect (see Figure~\ref{fig:cwm_domain_correct_graph_incorrect}).
Lastly, we include another example of correct domain witness prediction, with the goal of showing how the natural language reasoning is align and reflect the final answer (Figure~\ref{fig:cwm_reasoning}).

\label{appendix:model_witness_responses}


\begin{figure*}[t]
\centering
\fbox{%
  \begin{minipage}{1.0\textwidth}
    \small

    \textbf{Program Termination Analysis Prompt}

    You are an expert code analyzer specializing in program termination analysis. You will be given a C program to analyse. Every line of the program will begin with a line number (e.g., \texttt{13: x = x + 1} for line 13), which is for tracking and does not affect compilation. The program may either terminate for all inputs or diverge for some inputs. A function called \texttt{\_\_VERIFIER\_nondet\_X()} can return an arbitrary value of type \texttt{X}.

    Determine whether the program terminates on all inputs. Your goal is to earn the maximum number of points by answering correctly.

    \begin{compactitem}
      \item If the program always terminates, respond with:
        \begin{quote}
        \texttt{
\{\\
\ \ "verdict": true\\
\}
        }
        \end{quote}
        If you are correct, you will earn points.

      \item If the program can diverge (i.e., may not terminate for some inputs), respond with:
        \begin{quote}
        \texttt{
\{\\
\ \ "verdict": false,\\
\ \ "witness": \{ ... \}\\
\}
        }
        \end{quote}
        and provide a witness demonstrating the control flow graph of the non-termination loop. If you are correct and the witness is valid, you will earn points.

      \item If your answer is incorrect, you will lose points. The most severe error is predicting termination (\texttt{"verdict": true}) when the correct answer is non-termination (\texttt{"verdict": false}), which will result in the largest point deduction.

      \item If you cannot confidently determine whether the program always terminates, it is recommended to respond with:
        \begin{quote}
        \texttt{
\{\\
\ \ "verdict": null\\
\}
        }
        \end{quote}
        This will not earn points but also incurs no penalty.
    \end{compactitem}

    \textbf{Witness Generation for Non-Terminating Programs}

    When the verdict is false (non-terminating program), you MUST include a witness object that represents the control flow graph focused on the infinite loop structure.

    \textbf{JSON Structure (SV-COMP Witness Format):}
    \begin{compactitem}
      \item \textbf{Nodes:} \texttt{id}, \texttt{entry}, \texttt{cyclehead}
      \item \textbf{Edges:} \texttt{id}, \texttt{source}, \texttt{target}, \texttt{line}, \texttt{sourcecode}, \texttt{control}, \texttt{assumption}, \texttt{enterLoopHead}, \texttt{enterFunction}, \texttt{returnFrom}
    \end{compactitem}

    \textbf{Critical Constraints:}
    \begin{compactenum}
      \item Unique edge IDs and line numbers
      \item Loop head entries marked with \texttt{enterLoopHead: true}
      \item Node conventions: N1 is entry, N0 is cyclehead
      \item Cycle structure: last edge targets N0
      \item Focus on loop, no exit paths
      \item Graph flow: N1 $\rightarrow$ setup $\rightarrow$ N0 $\rightarrow$ loop body $\rightarrow$ N0 (cycle)
    \end{compactenum}

    \textbf{Graph Flow Structure:}
    \begin{compactenum}
      \item Start at N1 (entry node)
      \item Progress through necessary setup
      \item Enter the loop head N0
      \item Execute loop body
      \item Final edge targets N0, creating the infinite cycle
    \end{compactenum}

  \end{minipage}
}
\caption{Task requirements for program termination analysis in \svcomp format.}
\label{fig:termination_prompt}
\end{figure*}

\begin{figure*}[t]
\centering
\fbox{%
  \begin{minipage}{1.0\textwidth}
    \small

    \textbf{Examples:}
    \textbf{Example 1: Non-Terminating Program}
    \begin{quote}
    \texttt{%
1: extern int \_\_VERIFIER\_nondet\_int(void);\\
2: int main() \{\\
3: \ \ \ int x;\\
4: \ \ \ x = \_\_VERIFIER\_nondet\_int();\\
5:\\
6: \ \ \ while (x \% 2 == 0) \{\\
7: \ \ \ \ \ \ x = x + 2;\\
8: \ \ \}\\
9: \ \ \ return 0;\\
10: \}
    }
    \end{quote}

    \textbf{Expected answer:}
    \begin{quote}
    \texttt{%
\{\\
\ \ "verdict": false,\\
\ \ "witness": \{\\
\ \ \ \ "nodes": [\\
\ \ \ \ \ \ \{ "id": "N1", "entry": "true" \},\\
\ \ \ \ \ \ \{ "id": "N2" \},\\
\ \ \ \ \ \ \{ "id": "N0", "cyclehead": "true" \},\\
\ \ \ \ \ \ \{ "id": "N3" \}\\
\ \ \ \ ],\\
\ \ \ \ "edges": [\\
\ \ \ \ \ \ \{ "id": "E0", "source": "N1", "target": "N2", "line": 3, "sourcecode": "int x;" \},\\
\ \ \ \ \ \ \{ "id": "E1", "source": "N2", "target": "N0", "line": 4, "enterLoopHead": true, "sourcecode": "x = \_\_VERIFIER\_nondet\_int()" \},\\
\ \ \ \ \ \ \{ "id": "E2", "source": "N0", "target": "N3", "line": 6, "control": "condition-true", "assumption": "x \% 2 == 0", "sourcecode": "while (x \% 2 == 0) \{" \},\\
\ \ \ \ \ \ \{ "id": "E3", "source": "N3", "target": "N0", "line": 7, "enterLoopHead": true, "sourcecode": "x = x + 2" \}\\
\ \ \ \ ]\\
\ \ \}\\
\}
    }
    \end{quote}

    \textbf{Example 2: Terminating Program}
    \begin{quote}
    \texttt{%
1: extern int \_\_VERIFIER\_nondet\_int(void);\\
2: int main() \{\\
3: \ \ \ int i, j;\\
4: \ \ \ j = 0;\\
5: \ \ \ for (i = 0; i < 10; i++) \{\\
6: \ \ \ \ \ \ j = j + 1;\\
7: \ \ \}\\
8: \ \ \ return 0;\\
9: \}
    }
    \end{quote}

    \textbf{Expected answer:}
    \begin{quote}
    \texttt{%
\{\\
\ \ "verdict": true\\
\}
    }
    \end{quote}

  \end{minipage}
}
\caption{Few-shot examples for program termination and non-termination for the \svcomp prompt.}
\label{fig:termination_few_shot}
\end{figure*}

\begin{table*}[t]
    \centering
    \small
    \setlength{\tabcolsep}{8pt} 
    \begin{tabular}{l | c | c | c | c | c | c}
    \toprule
    \textbf{Model} & \textbf{FN} & \textbf{FP} & \textbf{Unknown} & \textbf{TP\_Unconfirmed\_Witness} & \textbf{TP} & \textbf{TN} \\
    \midrule
    GPT-5 & 13,492 & 6,607 & 6,658 & 10,356 & 4,874 & 4,422 \\
    Claude Sonnet-4.5 & 6,802 & 6,769 & 16,144 & 10,561 & 4,602 & 4,367 \\
    CWM & 11,672 & 11,534 & 6,504 & 9,449 & 4,717 & 4,184 \\ 
    Qwen3-30B & 7,040 & 6,451 & 6,543 & 10,221 & 4,387 & 4,529 \\ 
    GPT-4o & 3,887 & 1,595 & 11,684 & 13,160 & 3,924 & 1,334 \\
    \midrule
    Average & 8,579 & 6,591 & 9,507 & 10,749 & 4,501 & 3,767 \\  
    \bottomrule
    \end{tabular}
    \caption{
    Mean input code length (tokens) per score type, aggregated across all models. Incorrect predictions (FN, FP) and unknowns are associated with longer code inputs, while correct predictions (TP, TN) occur on shorter inputs. Unconfirmed witness cases show the highest average code length, suggesting models often abstain or struggle with validation on complex code.
    }
    \label{tab:mean_num_tokens_score_type}
\end{table*}

\begin{figure*}[t]
\centering
\fbox{%
  \begin{minipage}{1.0\textwidth}
    \small

    \textbf{Domain Witness Prompt}
    
    You are given a main function in C programming language that contains a loop or recursion and does not terminate for all inputs. \\

    Note: \_\_VERIFIER\_nondet\_\* defines a non-deterministic value that can be used to generate a random input. \\

    \textbf{Your task:} Predict for which non-deterministic inputs it will not terminate (diverge). \\

    \textbf{To achieve this:} Simulate the execution of the main function step-by-step. \\

    During simulation, detect early signs of divergence based on the logic of the loop. \\

    For example: if a variable used in the loop condition or update becomes positive while another grows unboundedly, or if the loop body increases \\
    a value that keeps the loop condition true indefinitely, this may indicate divergence. \\

    \textbf{Important:} \\
    Do not add arbitrary iteration limits unless strictly necessary for safety – the detection should be logical and correct. \\

    Assume the function uses only basic arithmetic and comparisons; no external state or side effects. \\

    \textbf{In your answer:} Write an expression involving the non-deterministic variables, which describes the conditions under which the function main is guaranteed to diverge. \\

    Please write an expression involving the non-deterministic variables and the logic gates ("and", "or", "not", and so on) for the range of input values causing the program to diverge. \\

    \textbf{An example would be as follows:} \\

\begin{quote}
\texttt{%
typedef enum \{false,true\} bool;\\
extern int \_\_VERIFIER\_nondet\_int(void);\\
int main() \{\\
\ \ \ int i;\\
\ \ \ i = \_\_VERIFIER\_nondet\_int();\\
\ \ \ while (i != 0) \{\\
\ \ \ \ \ \ if (i < 0) \{\\
\ \ \ \ \ \ \ \ i = i+2;\\
\ \ \ \ \ \ \ \ if (i < 0) \{\\
\ \ \ \ \ \ \ \ \ \ if (i <= -2147483648) return 0;\\
\ \ \ \ \ \ \ \ \ \ i = i*(-1);\\
\ \ \ \ \ \ \ \ \}\\
\ \ \ \ \ \ \}\\
\ \ \ \ \ \} else \{\\
\ \ \ \ \ \ \ \ i = i-2;\\
\ \ \ \ \ \ \ \ if (i > 0) \{\\
\ \ \ \ \ \ \ \ \ \ i = i*(-1);\\
\ \ \ \ \ \ \ \ \}\\
\ \ \ \ \ \ \}\\
\ \ \ \}\\
\ \ \ return 0;\\
\}\\
}
\end{quote}

\textbf{Expected answer:}\\
\((i \bmod 2 \ne 0)\) and \((i \ge -2147483649)\)

  \end{minipage}
}
\caption{A one-shot example prompt given to the model for the domain witness prompt.}
\label{fig:domain_witness_prompt}
\end{figure*}

\begin{figure*}[t]
\centering
\fbox{%
  \begin{minipage}{1.0\textwidth}
    \small
    \textbf{Input $C$ code:}

    \begin{quote}
    \texttt{%
1: // \\
2: //\\
3: // \\
4: \\
5: /*\\
6:  *  \\
7:  *  \\
8: */\\
9: \\
10: extern int \_\_VERIFIER\_nondet\_int(void);\\
11: \\
12: int main() \{\\
13: \ \ \ int a, x, y;\\
14: \ \ \ x = \_\_VERIFIER\_nondet\_int();\\
15: \ \ \ y = \_\_VERIFIER\_nondet\_int();\\
16: \ \ \ a = 0;\\
17: \ \ \ while (y > 0) \{\\
18: \ \ \ \ \ \ if (x < 0) \{\\
19: \ \ \ \ \ \ \ \ y = \~x;\\
20: \ \ \ \ \ \ \} else \{\\
21: \ \ \ \ \ \ \ \ y = y - 1;\\
22: \ \ \ \ \ \ \}\\
23: \ \ \ \}\\
24: \ \ \ return 0;\\
25: \}
    }
    \end{quote}

    \textbf{A correct domain witness:}
    \begin{quote}
    \texttt{x < -1 and y > 0}
    \end{quote}

    \textbf{A correct witness automaton prediction:} 

\begin{quote}
\texttt{%
\{ \\ 
"verdict": false,\\
"witness": \{\\
\ \ "nodes": [\\
\ \ \ \ \{ "id": "N1", "entry": "true" \},\\
\ \ \ \ \{ "id": "N2" \},\\
\ \ \ \ \{ "id": "N3" \},\\
\ \ \ \ \{ "id": "N4" \},\\
\ \ \ \ \{ "id": "N5" \},\\
\ \ \ \ \{ "id": "N0", "cyclehead": "true" \},\\
\ \ \ \ \{ "id": "N6" \}\\
\ \ ],\\
\ \ "edges": [\\
\ \ \ \ \{ "id": "E0", "source": "N1", "target": "N2", "line": 13, "sourcecode": "int a, x, y;" \},\\
\ \ \ \ \{ "id": "E1", "source": "N2", "target": "N3", "line": 14, "sourcecode": "x = \_\_VERIFIER\_nondet\_int()" \},\\
\ \ \ \ \{ "id": "E2", "source": "N3", "target": "N4", "line": 15, "sourcecode": "y = \_\_VERIFIER\_nondet\_int()" \},\\
\ \ \ \ \{ "id": "E3", "source": "N4", "target": "N5", "line": 16, "sourcecode": "a = 0" \},\\
\ \ \ \ \{ "id": "E4", "source": "N5", "target": "N0", "line": 17, "enterLoopHead": true, "sourcecode": "while (y > 0){" \},\\
\ \ \ \ \{ "id": "E5", "source": "N0", "target": "N6", "line": 18, "control": "condition-true", "assumption": "x < 0", "sourcecode": "if (x < 0 )" \},\\
\ \ \ \ \{ "id": "E6", "source": "N6", "target": "N0", "line": 19, "enterLoopHead": true, "sourcecode": "y = \~x" \}\\
\ \ ]
\} 
\}
} 
} 
\end{quote}

  \end{minipage}
}

\caption{
A sample where CWM produces both a \textbf{correct domain witness prediction} and a \textbf{correct graph-based prediction} (\svcomp). The model's output includes natural language reasoning (omitted for brevity) and a JSON answer: the verdict is \texttt{false} for NT and \texttt{true} for T, with a witness provided for NT cases.
}
\label{fig:cmw_domain_correct_graph_correct}
\end{figure*}

\begin{figure*}[t]
\centering
\fbox{%
  \begin{minipage}{1.0\textwidth}
    \small
    \textbf{Input $C$ code:}

    \begin{quote}
    \texttt{%
1: typedef enum \{false,true\} bool;\\
2: \\
3: extern int \_\_VERIFIER\_nondet\_int(void);\\
4: \\
5: int main() \{\\
6: \ \ \ \ int i;\\
7: \ \ \ \ i = \_\_VERIFIER\_nondet\_int();\\
8: \ \ \ \ \\
9: \ \ \ \ while (i >= -5 \&\& i <= 5) \{\\
10: \ \ \ \ \ \ \ \ if (i > 0) \{\\
11: \ \ \ \ \ \ \ \ \ \ \ \ i = i-1;\\
12: \ \ \ \ \ \ \ \ \} \\
13: \ \ \ \ \ \ \ \ if (i < 0) \{\\
14: \ \ \ \ \ \ \ \ \ \ \ \ i = i+1;\\
15: \ \ \ \ \ \ \ \ \} \\
16: \ \ \ \ \} \\
17: \ \ \ \ \\
18: \ \ \ \ return 0;\\
19: \}
    }
\end{quote}

    \textbf{A wrong domain witness:}
    \begin{quote}
    \texttt{i = 0}
    \end{quote}

    \textbf{The correct domain witness:}
    \begin{quote}
    \texttt{i >= -5 and i <= 5}
    \end{quote}

    \textbf{A correct witness automaton prediction:} 

\begin{quote}
    \texttt{%
\{ \\
  "verdict": false,\\
  "witness": \{\\
    "nodes": [\\
      \{ "id": "N1", "entry": "true" \},\\
      \{ "id": "N2" \},\\
      \{ "id": "N0", "cyclehead": "true" \}\\
    ],\\
    "edges": [\\
      \{ "id": "E0", "source": "N1", "target": "N2", "line": 6, "sourcecode": "int i;" \},\\
      \{ "id": "E1", "source": "N2", "target": "N0", "line": 7, "enterLoopHead": true, "sourcecode": "i = \_\_VERIFIER\_nondet\_int()" \},\\
      \{ "id": "E2", "source": "N0", "target": "N0", "line": 9, "control": "condition-true", "assumption": "i == 0", "enterLoopHead": true, "sourcecode": "while (i >= -5 \&\& i <= 5) {" \}\\
    ]\\
  \}\\
\}\\
    }
    }
\end{quote}

  \end{minipage}
}

\caption{
A sample where CWM produces a \textbf{wrong domain witness prediction} but a \textbf{correct graph-based prediction} (\svcomp). The model's output includes natural language reasoning (omitted for brevity) and a JSON answer: the verdict is \texttt{false} for NT and \texttt{true} for T, with a witness provided for NT cases. The witness automaton graph demonstrates a non-terminating loop: the variable $i$ may start at any value from $-5$ to $5$, and after several iterations, $i$ eventually reaches $0$ (as indicated by the assumption property in edge E2), causing the loop to continue indefinitely. So the correct domain is $i >= -5$ and $i <= 5$.
}
\label{fig:cwm_domain_incorrect_graph_correct}
\end{figure*}

\begin{figure*}[t]
\centering
\fbox{%
  \begin{minipage}{1.0\textwidth}
    \small
    \textbf{Input $C$ code:}
    \begin{quote}
\texttt{%
1: typedef enum \{false,true\} bool;\\
2: \\
3: extern int \_\_VERIFIER\_nondet\_int(void);\\
4: \\
5: int main() \{\\
6: \ \ \ \ int i;\\
7: \ \ \ \ i = \_\_VERIFIER\_nondet\_int();\\
8: \\
9: \ \ \ \ while (i < 0) \{\\
10: \ \ \ \ \ \ \ \ if (i != -5) \{\\
11: \ \ \ \ \ \ \ \ \ \ \ \ i = i+1;\\
12: \ \ \ \ \ \ \ \ \} \\
13: \ \ \ \ \} \\
14: \\
15: \ \ \ \ return 0;\\
16: \}
}
\end{quote}

    \textbf{A correct domain witness prediction:}
    \begin{quote}
    \texttt{i <= -5}
    \end{quote}

    \textbf{A wrong witness automaton prediction:} 
\begin{quote}
\texttt{%
\{ \\
\ \ "verdict": false,\\
\ \ "witness": \{\\
\ \ \ \ "nodes": [\\
\ \ \ \ \ \ \{"id": "N1", "entry": "true"\},\\
\ \ \ \ \ \ \{"id": "N2"\},
\ \ \ \ \ \ \{"id": "N0", "cyclehead": "true"\},\\
\ \ \ \ \ \ \{"id": "N3"\}
\ \ \ \ ],\\
\ \ \ \ "edges": [\\
\ \ \ \ \ \ \{"id": "E0", "source": "N1", "target": "N2", "line": 6, "sourcecode": "int i;"\},\\
\ \ \ \ \ \ \{"id": "E1", "source": "N2", "target": "N0", "line": 7, "enterLoopHead": true, "sourcecode": "i = \_\_VERIFIER\_nondet\_int()"\},\\
\ \ \ \ \ \ \{"id": "E2", "source": "N0", "target": "N3", "line": 9, "control": "condition-true", "assumption": "i < 0", "sourcecode": "while (i < 0) {"\},\\
\ \ \ \ \ \ \{"id": "E3", "source": "N3", "target": "N0", "line": 11, "enterLoopHead": true, "sourcecode": "i = i+1"\}\\
\ \ \ \ ]
\ \ \}
\}
}
}
\end{quote}
    \textbf{A correct witness automaton prediction:} 
\begin{quote}
\texttt{%
\{ \\
\ \ "verdict": false,\\
\ \ "witness": \{\\
\ \ \ \ "nodes": [\\
\ \ \ \ \ \ \{"id": "N1", "entry": "true"\},\\
\ \ \ \ \ \ \{"id": "N2"\},\\
\ \ \ \ \ \ \{"id": "N0", "cyclehead": "true"\},\\
\ \ \ \ \ \ \{"id": "N3"\}\\
\ \ \ \ ],\\
\ \ \ \ "edges": [\\
\ \ \ \ \ \ \{"id": "E0", "source": "N1", "target": "N2", "line": 6, "sourcecode": "int i;"\},\\
\ \ \ \ \ \ \{"id": "E1", "source": "N2", "target": "N0", "line": 7, "enterLoopHead": true, "sourcecode": "i = \_\_VERIFIER\_nondet\_int()"\},\\
\ \ \ \ \ \ \{"id": "E2", "source": "N0", "target": "N3", "line": 9, "control": "condition-true", "assumption": "i < 0", "sourcecode": "while (i < 0) {"\},\\
\ \ \ \ \ \ \{"id": "E3", "source": "N3", "target": "N0", "line": 13, "enterLoopHead": true, "sourcecode": "}"\}
\ \ \ \ ]
\ \ \}
\}
}
\end{quote}

  \end{minipage}
}
\caption{
A sample where CWM produces a \textbf{correct domain witness prediction} but a \textbf{incorrect graph-based prediction} (\svcomp). The model's output includes natural language reasoning (omitted for brevity) and a JSON verdict (true for T, false for NT), with a witness provided for NT cases. Among 10 predictions, a correct witness automaton prediction is also found and shown below the incorrect one.
}
\label{fig:cwm_domain_correct_graph_incorrect}
\end{figure*}

\begin{figure*}[t]
\centering
\fbox{%
  \begin{minipage}{1.0\textwidth}
    \small
    \textbf{Input $C$ code:}

       \begin{quote}
\texttt{%
1: typedef enum \{false,true\} bool;\\
2: \\
3: extern int \_\_VERIFIER\_nondet\_int(void);\\
4: \\
5: int main() \{\\
6: \ \ \ \ int i;\\
7: \ \ \ \ i = \_\_VERIFIER\_nondet\_int();\\
8: \\
9: \ \ \ \ while (i < 0) \{\\
10: \ \ \ \ \ \ \ \ if (i != -5) \{\\
11: \ \ \ \ \ \ \ \ \ \ \ \ i = i+1;\\
12: \ \ \ \ \ \ \ \ \} \\
13: \ \ \ \ \} \\
14: \\
15: \ \ \ \ return 0;\\
16: \}
}
\end{quote}

    \textbf{Reasoning and answer:}
    Okay, so I need to figure out for which values of the non-deterministic variable `i` the given C program will not terminate. Let me look at the code again.The program starts by assigning `i` a non-deterministic integer value. Then it enters a while loop that continues as long as `i` is less than 0. Inside the loop, there's an if statement that checks if `i` is not equal to -5. If that's true, it increments `i` by 1. If `i` is -5, the loop body does nothing, and the loop condition is checked again.Hmm, so let's think about what happens in different scenarios.First, if `i` starts at a value less than -5, say -10. Then, in each iteration, since `i` is not -5, it gets incremented by 1. So it would go from -10 to -9, then -8, and so on, until it reaches -5. At that point, when `i` is -5, the if condition `i != -5` becomes false, so the increment doesn't happen. The loop condition `i < 0` is still true because -5 is less than 0, so the loop continues. But now, since `i` is -5, the if statement doesn't execute, so `i` remains -5. This means the loop will keep checking the condition `i < 0`, which is true, and since `i` doesn't change, it will loop indefinitely. So any starting value of `i` that is less than or equal to -5 will cause the program to diverge.Wait, but what if `i` starts at a value greater than or equal to 0? Then the loop condition `i < 0` is false from the start, so the loop doesn't execute at all, and the program terminates immediately. That's not a problem.What about if `i` starts between -5 and 0, like -3? Then, in the first iteration, since `i` is not -5, it gets incremented by 1, becoming -2. Then the loop condition is checked again. Since -2 is still less than 0, the loop continues. Again, `i` is not -5, so it increments to -1. Then the loop condition is checked again. -1 is less than 0, so loop again. Increment to 0. Now, the loop condition `i < 0` is false, so the loop exits, and the program terminates. So starting values between -4 and -1 (inclusive) will eventually reach 0 and terminate.But wait, what if `i` starts at -5? Then, in the first iteration, the if condition `i != -5` is false, so `i` doesn't change. The loop condition `i < 0` is true, so the loop continues. But since `i` remains -5, this will loop forever. So starting at -5 is a problem.Similarly, if `i` starts at any value less than -5, like -6, -7, etc., it will increment by 1 each time until it reaches -5. Once it reaches -5, the increment stops, and the loop continues indefinitely because `i` is still less than 0. So any starting value less than or equal to -5 will cause divergence.Wait, but let's test this with an example. Suppose `i` starts at -6. First iteration: `i` is -6 (not -5), so increment to -5. Now, `i` is -5. Next iteration: `i` is -5, so the if condition is false, no increment. Loop condition `i < 0` is true, so loop again. `i` remains -5. This repeats forever. So yes, starting at -6 leads to divergence.Another example: `i` starts at -5. First iteration: `i` is -5, so no increment. Loop condition is true, loop again. `i` remains -5. Diverges.If `i` starts at -4: first iteration, `i` is -4 (not -5), increment to -3. Loop condition true. Next iteration, `i` is -3, increment to -2. Loop condition true. Next, `i` is -2, increment to -1. Loop condition true. Next, `i` is -1, increment to 0. Loop condition false, exit. Terminates.So the divergence occurs when the initial value of `i` is less than or equal to -5. Because in those cases, the loop will eventually reach -5 (if starting below -5) or start at -5, and then get stuck in an infinite loop.Wait, but what if `i` starts at a value less than -5, but when it increments, it skips over -5? No, because it increments by 1 each time. So if `i` starts at -6, it goes to -5, then stops incrementing. If it starts at -1000, it will increment 995 times to reach -5, then stop. So any starting value less than or equal to -5 will eventually reach -5 and then loop forever.Therefore, the condition for divergence is `i <= -5`. But wait, let me check the code again. The loop condition is `i < 0`. So when `i` is -5, the loop continues. But when `i` is -5, the if statement `i != -5` is false, so `i` doesn't change. So the loop will run indefinitely.Therefore, the program diverges if and only if the initial value of `i` is less than or equal to -5.But wait, what if `i` is exactly -5? Then the loop runs once, checks `i < 0` (true), enters the loop body. Since `i` is -5, the if condition is false, so `i` remains -5. Then the loop condition is checked again, still true, and this repeats forever. So yes, `i = -5` is a diverging case.Therefore, the divergence condition is `i <= -5`.But let me think again. Suppose `i` is -5. The loop runs, but `i` doesn't change. So it's an infinite loop. If `i` is less than -5, like -6, then it increments to -5, then gets stuck. So any `i <= -5` will diverge.Therefore, the answer should be `i <= -5`.Wait, but the problem says to write an expression involving the non-deterministic variables and logic gates. So in this case, the non-deterministic variable is `i`, and the condition is `i <= -5`.But let me check if there's any other possibility. For example, what if `i` is a very large negative number, like -2147483648 (assuming 32-bit int). Then, incrementing by 1 each time until it reaches -5. But since integers in C can overflow, but in this case, since we're dealing with negative numbers and incrementing, it should be okay. Wait, but if `i` is -2147483648 (the minimum value for a 32-bit signed int), then incrementing it by 1 would result in -2147483647, and so on, until it reaches -5. So even in that case, it would eventually reach -5 and then loop forever. So the condition `i <= -5` holds.Therefore, the program diverges if and only if the initial value of `i` is less than or equal to -5.So the answer should be `i <= -5`.Wait, but let me make sure there aren't any other edge cases. For example, what if `i` is 0? Then the loop doesn't run, program terminates. If `i` is positive, same thing. If `i` is between -4 and -1, it increments until it reaches 0 and exits. So yes, only `i <= -5` causes divergence. Therefore, the condition is `i <= -5`.</think>\textbf{<answer>i <= -5</answer>}

  \end{minipage}
}
\caption{ A correct domain witness prediction by CWM: reasoning followed by the logical answer, $i \le -5$.}
\label{fig:cwm_reasoning}
\end{figure*}


\begin{table*}[t!]
    \centering
    \footnotesize
    \setlength{\tabcolsep}{10pt}
    \begin{tabular}{l | p{10cm} | c }
        \toprule
        \textbf{Subcategory} & \textbf{Description} & \textbf{Samples} \\
        \midrule
        BitVectors & Programs involving bit-precise arithmetic operations. & $34$ \\
        MainControlFlow & Programs with complex control flow (e.g., loops, conditionals, recursion). & $281$ \\
        MainHeap & Programs that utilize dynamic heap memory allocation and manipulation. & $202$ \\
        Other & Miscellaneous benchmarks related to termination verification. & $1{,}822$ \\
        \midrule
        Total & Union over subcategories. We follow \svcomp\ 2025 and exclude $11$ invalid tasks ($9$ from ``MainControlFlow'' and $2$ from ``Other'') & $2{,}328$ \\
        \bottomrule
    \end{tabular}
    \caption{Distribution of termination verification tasks in  \svcomp 2025 across four subcategories.}
    \label{tab:termination_category_summary}
\end{table*}

\begin{table*}[t]
\centering
\small
\setlength{\tabcolsep}{5pt}
\begin{tabular}{@{}lcccccc@{}}
\toprule
& \multicolumn{3}{c}{\textbf{\svcomp-1}} & \multicolumn{3}{c}{\textbf{\svcomp-TTS@10}} \\
\cmidrule(lr){2-4} \cmidrule(lr){5-7}
\textbf{Method} & \textbf{Score} & \textbf{F1 (T)} & \textbf{F1 (NT)} & \textbf{Score} & \textbf{F1 (T)} & \textbf{F1 (NT)} \\
\midrule
GPT-5             & \textbf{3,211} ($\pm$ 185) & \textbf{.98} ($\pm$ .001) & \textbf{.97} ($\pm$ .002) & \textbf{3,520} ($\pm$ 47)  & \textbf{.98} ($\pm$ .001) & \textbf{.96} ($\pm$ .002) \\
Claude Sonnet-4.5 & 2,737 ($\pm$ 184)          & \textbf{.98} ($\pm$ .002) & .96 ($\pm$ .003)          & 3,448 ($\pm$ 49)           & .94 ($\pm$ .003)          & .89 ($\pm$ .005) \\
CWM               & 1,491 ($\pm$ 324)          & .91 ($\pm$ .004)          & .80 ($\pm$ .011)          & 3,129 ($\pm$ 80)           & .84 ($\pm$ .003)          & .63 ($\pm$ .008) \\
Qwen3-30B         & 225 ($\pm$ 214)            & .90 ($\pm$ .004)          & .78 ($\pm$ .010)          & 2,895 ($\pm$ 84)           & .83 ($\pm$ .003)          & .59 ($\pm$ .005) \\
GPT-4o            & -5,145 ($\pm$ 750)         & .73 ($\pm$ .004)          & .49 ($\pm$ .006)          & 546 ($\pm$ 146)            & .73 ($\pm$ .004)          & .50 ($\pm$ .006) \\
\midrule
PROTON            & \textbf{3,685}             & --                         & --                         & --                         & --                         & -- \\
UAutomizer        & 3,334                      & --                         & --                         & --                         & --                         & -- \\
AProVE            & 2,219                      & --                         & --                         & --                         & --                         & -- \\
\midrule
Max               & 4,079                      & 1.0                        & 1.0                        & 4,079                      & 1.0                        & 1.0 \\
Min               & -50,064                    & 0.0                        & 0.0                        & -50,064                    & 0.0                        & 0.0 \\
\bottomrule
\end{tabular}

\vspace{2pt}
\caption{\svcomp-1, \svcomp-TTS@10, and F1 scores over termination (T) and non-termination (NT). Results are reported as mean $\pm$ standard deviation over 100 bootstrap runs; \texttt{unknown} predictions are treated as errors for F1.}
\label{tab:model_performance_detailed}
\vspace{2pt}
\end{table*}


\end{document}